\documentclass[final,5p,times,twocolumn]{elsarticle}

\usepackage[T1]{fontenc}
\usepackage[utf8]{inputenc}
\usepackage{amsmath,amssymb}
\usepackage{booktabs,tabularx,array,multirow}
\usepackage{float}
\usepackage{xcolor}
\usepackage[most]{tcolorbox}
\usepackage{tikz}
\usetikzlibrary{arrows.meta,positioning,shapes.geometric,fit,backgrounds}
\usepackage{graphicx}
\usepackage{url}
\usepackage[hidelinks]{hyperref}
\usepackage[expansion=false]{microtype}
\tcbset{colback=gray!4, colframe=black!20, boxsep=2pt,
        arc=1mm, left=2pt, right=2pt, top=2pt, bottom=2pt,
        fonttitle=\small\bfseries}
\newtcolorbox{conceptbox}[1][]{%
  title={Box~1: Conceptual Glossary: Goal, Intention, Subgoal},#1}
\newtcolorbox{metricbox}[1][]{%
  title={Box~2: Metric and Protocol Definitions},#1}
\newtcolorbox{protocolbox}[1][]{%
  title={Box~3: Reporting Guidelines},#1}

\newcolumntype{Y}{>{\raggedright\arraybackslash}X}
\newcolumntype{P}[1]{>{\raggedright\arraybackslash}p{#1}}

\begin{document}

\begin{frontmatter}

\title{Language-Augmented Video Action Anticipation:\\
Design Fundamentals, Benchmarks, and Open Challenges}


\author[aff1]{Mahsa Mohammadi\corref{cor1}}
\ead{mm1510@exeter.ac.uk}
\cortext[cor1]{Corresponding author.}
\author[aff1]{Zeyu Fu}
\author[aff1]{Sareh Rowlands}
\affiliation[aff1]{%
  organization={Department of Computer Science, University of Exeter},
  addressline={Harrison Building, Streatham Campus, North Park Road},
  city={Exeter}, postcode={EX4 4QF},
  state={Devon}, country={United Kingdom}}

\begin{abstract}
Action anticipation predicts future human actions from partial video
under incomplete context and temporal uncertainty. Recent systems
introduce large language models (LLMs), vision--language models
(VLMs), or language-derived semantics at different stages, but reported
gains are difficult to interpret when task formulation, visual
pretraining, supervision, decoder design, and evaluation code change
simultaneously.

The central contribution of this review is an evidence-aware design map
that crosses task regime with the point at which language-derived
information intervenes. We characterise task regimes along six axes. These axes organise the
literature into five broad task families: single-action, sequence,
object-interaction, cross-view, and planning-oriented settings. C1--C3
locate interventions in context construction, goal/intention modelling,
and future decoding, while C4 is treated as an adjacent, emerging
grounding/executability extension. Unlike a generic processing pipeline,
the map links each intervention to an appropriate counterfactual,
failure diagnosis, and permissible evidence claim.

Supporting contributions include a protocol-level audit of Ego4D-LTA
and EPIC-KITCHENS-100, a multidimensional evidence profile, and the
Backbone-Aware Comparison and Ablation Protocol (BCAP). The unresolved
EK-100 record is treated as a reporting-comparability case study and is
not used as a leaderboard. Evidence for LLM benefits, goal ambiguity, and
horizon effects is therefore formulated as testable hypotheses requiring
matched validation, not as causal conclusions. The accompanying package contains the coded evidence, source locators,
protocol metadata, and versioned catalogue used in the review.
\end{abstract}

\begin{keyword}
Action anticipation \sep Long-term forecasting \sep
Large language models \sep Vision--language models \sep
Protocol comparability \sep Ego4D \sep EPIC-KITCHENS
\end{keyword}


\end{frontmatter}

\section{Introduction}
\label{sec:intro}

\subsection{Why Anticipation Differs from Recognition}
\label{sec:challenge}

Action recognition on established benchmarks such as
Kinetics-700~\citep{smaira2020k700} and
UCF-101~\citep{soomro2012ucf101} is now a comparatively mature
problem, with steady year-on-year gains from large-scale pretraining.
Anticipating a future action before it begins, however, is not simply
a more difficult variant of recognition; it is a qualitatively
different inference problem.

Recognition is grounded in observed evidence, whereas anticipation
must infer an \emph{unobserved} future from partial and often
underspecified context. This difference introduces at least three
structural challenges.

\textbf{(i)~Goal ambiguity.} The same visible behaviour may admit
multiple plausible continuations. For example, reaching toward a
kitchen counter may precede picking up a knife, a cup, or a phone.
Without some representation of the actor's underlying
\emph{goal}, the space of future actions remains weakly
constrained, even when past observations are encoded accurately.

\textbf{(ii)~Temporal scale mismatch.} Human activity unfolds over
multiple temporal scales, from fine-grained motor gestures to
short sub-tasks and longer procedural objectives. Models trained
primarily on short clips often fail to preserve the long-range
context required to disambiguate these futures, especially when
the relevant intent is expressed only over extended horizons.

\textbf{(iii)~World knowledge and executability.} Plausible future
actions must be statistically likely, physically feasible, semantically
coherent, and consistent with the actor's broader objective. Predicting that \texttt{crack\_egg} may follow
\texttt{take\_bowl} can be learned as a recurrent action pattern;
inferring that it is a sensible next step because the actor is
preparing an omelette requires a richer model of commonsense and
task structure. Explicit, broadly pretrained procedural priors and language-mediated
goal interfaces were uncommon in earlier anticipation systems, which
often relied more heavily on visual-temporal regularities.

These challenges explain why stronger visual recognition alone is
insufficient. Anticipation also requires models that integrate partial
perceptual evidence with long-range context, structured knowledge, and
reasoning about plausible future behaviour.

\subsection{Why LLMs Entered This Space}
\label{sec:llm-entry}

To appreciate why LLMs entered action anticipation, it helps to
contrast two eras of design. In the pre-LLM era, roughly 2010 to
2021, methods relied on visual-temporal regularities: hand-crafted
trajectories~\citep{wang2013idt}, learned spatiotemporal
features~\citep{tran2015c3d,carreira2017quovadis}, and recurrent
sequence models~\citep{furnari2019rulstm}. Many widely used systems encoded the observed prefix into a
fixed-dimensional vector and decoded the next action through a softmax
head or LSTM unrolling. This did not make goal modelling or constraint
reasoning architecturally impossible: inverse-reinforcement-learning,
latent-intention, and structured prediction approaches already
provided counterexamples. What was uncommon was access to broadly
pretrained procedural priors and language-mediated interfaces that
could be reused without task-specific goal annotation. From 2022
onward, LLMs and VLMs expanded this design space; they did not create
an absolute capability boundary. LLM-augmented systems can condition
predictions on
inferred high-level goals, generate procedurally coherent futures
using broad commonsense priors, impose plausibility constraints
through language-grounded reasoning, and interface with external
symbolic structures such as action programs and knowledge graphs.

The growing use of large language models (LLMs) in action
anticipation is therefore largely driven by a good match between
their strengths and the structural difficulties outlined above. In
particular, LLMs provide access to broad procedural priors,
goal-level abstractions, and semantically structured future
generation when coupled with sufficiently strong visual context.

For \emph{goal ambiguity}, LLM-based methods can condition future
prediction on inferred high-level intentions while still using local
motion cues. AntGPT~\citep{zhao2024antgpt}
is representative in this respect: it formulates long-term action
anticipation through both bottom-up temporal completion and
top-down goal inference, showing that explicit reasoning over
latent goals can improve future-action prediction in long-horizon
settings. We note that such a decoupling of goal inference from
decoding is not architecturally impossible in a recurrent model: a
latent goal estimator feeding a goal-conditioned LSTM decoder is
readily expressible, and Mascar\'o et~al.~\citep{mascaro2023intention}
is precisely such a design. One important practical change
introduced by LLMs is access to broad procedural priors that can make
an inferred goal informative without task-specific goal supervision;
this change co-occurs with shifts in scale, optimisation, multimodal
pretraining, and interface design, so it should not be interpreted as
a controlled historical attribution.

For \emph{temporal scale mismatch}, LLM-style decoders can operate
on recognised action histories or other compact structured
summaries, making long-range temporal modelling more tractable
than relying on frame-level evidence alone. This is especially
useful when anticipation depends less on immediate appearance and
more on procedural coherence across multiple future steps.

For the \emph{world-knowledge gap}, language-augmented models can
impose commonsense and task-level constraints on predicted
futures. PlausiVL~\citep{mittal2024plausivl}, for example,
explicitly optimises for plausible action generation using
counterfactual temporal and verb--noun logical constraints,
together with a long-horizon repetition loss that discourages
degenerate forecasts of the kind that plagued earlier sequence
decoders.

At the same time, the benefit of LLM augmentation is not uniform
across all settings. On the short-horizon EPIC-KITCHENS-100
benchmark, the V-JEPA line reports strong results using large-scale
self-supervised visual pretraining and an attentive probe, with no
LLM in the anticipation
head~\citep{assran2025vjepa2,murLabadia2026vjepa21}. This motivates testing whether scaled visual pretraining reduces the
marginal advantage of LLM augmentation in short-horizon settings; the
current cross-paper evidence cannot establish this effect. We return to
the reporting incompatibility in Section~\ref{sec:benchmarks}.

The central question, therefore, is not whether LLMs help in the
abstract, but under what conditions and through which design
choices they add value beyond stronger visual backbones. Current evidence motivates the hypothesis that the value of
language-related components may depend on task regime, goal branching,
and prediction horizon, as well as on the need for explicit plausibility,
procedural knowledge, or constraint satisfaction.

\subsection{Gap in Existing Surveys}
\label{sec:gap}

Table~\ref{tab:survey-comparison} compares the present survey with
the most relevant prior reviews in adjacent areas.

\begin{table}[t]
\scriptsize
\centering
\caption{Comparison with related reviews
(\checkmark~=~yes; $\circ$~=~partial; no~=~not covered).
``Design-centric'' requires an explicit component-level analysis of
where language enters anticipation; ``protocol-aware'' requires an
audit of evaluation comparability, not benchmark coverage alone. Cells refer to the version listed in ``Ver.'' and are therefore
version-specific. Source locations and coding rationales are provided
in Supplementary Table~S0.}
\label{tab:survey-comparison}
\setlength{\tabcolsep}{1.4pt}
\renewcommand{\arraystretch}{1.15}
\begin{tabularx}{\columnwidth}{@{}lcccccc@{}}
\toprule
\textbf{Survey} & \textbf{Ver.}
  & \begin{tabular}[c]{@{}c@{}}LLM\\focused\end{tabular}
  & \begin{tabular}[c]{@{}c@{}}Anticip.\\focused\end{tabular}
  & \begin{tabular}[c]{@{}c@{}}Design\\centric\end{tabular}
  & \begin{tabular}[c]{@{}c@{}}Protocol\\aware\end{tabular}
  & \begin{tabular}[c]{@{}c@{}}LLM vs.\\backbone\end{tabular} \\
\midrule
Tang et al.~\citep{tang2025survey}
  & v1, 2025 & \checkmark & no & $\circ$ & no & no \\
Zhong et al.~\citep{zhong2023survey}
  & v2, 04/2026 & $\circ$ & \checkmark & no & $\circ$ & no \\
Kong \& Fu~\citep{kong2022survey}
  & 2022 & no & \checkmark & no & $\circ$ & no \\
Lai et al.~\citep{lai2024survey}
  & v1, 2024 & \checkmark & \checkmark & $\circ$ & $\circ$ & no \\
Hu et al.~\citep{hu2022survey}
  & 2022 & no & $\circ$ & no & $\circ$ & no \\
Stergiou \& Poppe~\citep{stergiou2025abouttime}
  & 2025 & $\circ$ & $\circ$ & no & $\circ$ & no \\
\textbf{This survey}
  & 2026 & \checkmark & \checkmark & \checkmark & \checkmark & \checkmark \\
\bottomrule
\end{tabularx}
\end{table}

The neighbouring literature has moved quickly, so the novelty claim is
intentionally narrow. Current reviews already cover language-model
methods, datasets, metrics, and performance comparisons in action
anticipation and video understanding~\citep{zhong2023survey,lai2024survey,tang2025survey,stergiou2025abouttime}.
We therefore do not claim the first survey of LLM-based anticipation.

The distinction is the combination of three elements. First, methods
are organised by the position at which language-derived information
intervenes in the anticipation factorisation, yielding the non-unique
C1--C4 analytical frame. Second, protocol comparability is audited explicitly instead of being
inferred from a shared metric label;
Section~\ref{sec:benchmarks} documents an unresolved short-horizon
reporting incompatibility. Third, anticipation is linked to grounding,
executability, efficiency, and streaming through a common evidence
frame and the BCAP reporting/ablation protocol. Table~\ref{tab:survey-comparison}
and Supplementary Table~S0 provide the version-specific evidence for
this positioning.

\subsection{Three Organising Questions}
\label{sec:questions}

Consistent with design-oriented survey structures in computer
vision~\citep{chang2026diffusion}, this review is organised around
three questions:

\begin{enumerate}
  \item \textbf{What functional components define
  language-augmented action anticipation systems?}

  \item \textbf{Which major design factors shape each component,
  and what trade-offs do they introduce?}

  \item \textbf{Under benchmarked anticipation protocols, what
  evidence distinguishes language-related contributions from
  backbone, pretraining, probe, and evaluation effects?}
\end{enumerate}

\subsection{Contributions and Organisation}
\label{sec:org}

This review has one central contribution and three supporting
contributions.

\textbf{Central contribution: an evidence-aware design map.}
We combine the task-regime taxonomy of Section~\ref{sec:task-regimes}
with a non-unique C1--C4 insertion-point analysis. The resulting map does
more than name successive modules: for each design intervention it
identifies the relevant counterfactual control, the expected failure
signature, and the strength of claim that the available evidence can
support. This makes it possible to separate, for example, a gain caused
by stronger C1 representation learning from one caused by C2 goal
conditioning, C3 sequence control, or an emerging C4 feasibility check.
Table~\ref{tab:central-design-map} operationalises this central contribution
by placing intervention, required counterfactual, characteristic failure,
and permissible claim in one view. The map is applied to 25 focal methods
within a retained synthesis set ($n{=}69$), with 10 additional evidence
records reported separately.
The focal set contains 16 core LLM/VLM-augmented methods, six non-LLM diagnostic/exemplar methods, and three planning/robotics
boundary cases; the
complete inventory and assignment sensitivity analysis are provided in
Supplementary Sections~S1 and~S6.

\textbf{Supporting contribution 1: protocol and claim audit.}
We audit reported Ego4D-LTA and EPIC-KITCHENS results at the level of
metric definition, split, version, horizon, evaluator traceability, and
row provenance. The analysis distinguishes source-described alignment,
code-verified alignment, and independent reproduction, and treats the
unresolved EK-100 discrepancy as a reporting incompatibility and does
not present it as a cross-family leaderboard.

\textbf{Supporting contribution 2: hypothesis-generating synthesis.}
We synthesise recurrent design patterns across task regimes while
separating reported system performance from causal attribution to a
language component. Claims about LLM benefit, goal ambiguity, and
prediction horizon are stated as well-motivated hypotheses whose
validation requires matched backbones, pretraining, supervision,
decoders, and evaluation code.

\textbf{Supporting contribution 3: BCAP and reusable artefacts.}
We propose the Backbone-Aware Comparison and Ablation Protocol (BCAP),
which separates descriptive native-system audit from causal
within-family ablation. BCAP-Swappable is a proposed validation protocol,
not an experiment executed in this review; the retrospective BCAP-Native
audit is an illustrative documentation exercise and does not validate
the full protocol empirically. The submission package includes the coded
evidence inventory, claim--evidence map, protocol metadata, source
locators, and a versioned ``Awesome Language-Augmented Action
Anticipation'' catalogue organised by task regime and C1--C4. Version
1.0.0 is publicly available in the \href{https://github.com/mahsa7290/language-augmented-action-anticipation}{GitHub repository} (release \texttt{v1.0.0}) and is also included with the submission.

The remainder defines the scope and methodology, introduces the task and
C1--C4 maps, organises the literature by comparative design findings,
audits benchmark comparability and deployment constraints, and concludes
with testable hypotheses, BCAP, open challenges, and limitations.

\subsection{Relation to Neural Computation and Learning Systems}
\label{sec:pr-fit}

Action anticipation is a neural computation problem involving partial
observations, learned representations, latent intentions, structured
sequence prediction, and optional feasibility constraints. Within this
view, C1 concerns representation learning, C2 latent-variable
inference, C3 neural sequence decoding, and C4 constrained prediction
or planning. LLMs and VLMs are parameterisations of these functions,
not substitutes for them. The central question--when language-derived
semantics add value beyond a strong representation learner--therefore
connects directly to attribution, calibration, reproducibility, and
compute in neural learning systems.

\section{Scope, Review Methodology, and Problem Definition}
\label{sec:scope}

\subsection{Scope and Task Definition}

This review focuses on video action anticipation, namely
the task of predicting one or more future actions from a partial
video observation under standardised evaluation protocols. The
predicted outputs may take the form of verb and noun labels,
action categories, or free-text procedural steps, depending on the
benchmark and modelling formulation.

The core tasks covered in this review are long-term action
anticipation, short-term object-interaction anticipation, and
procedure anticipation in both egocentric and third-person video
settings.

We also discuss several adjacent areas, including visual
planning, streaming video dialogue, and world-model-based
robotics, but only when they directly inform anticipation system
design or report results that are relevant to established
anticipation benchmarks. These neighbouring topics are therefore
used to clarify the broader design space. They are not reviewed
exhaustively in their own right.

\subsection{Review Methodology}
\label{sec:method}

\textbf{Retrieval and verification.}
The retained synthesis was assembled through three documented windows.
Windows~1 (1 January 2018--30 April 2025) and~2 (1 May 2025--30 April
2026) combined venue/arXiv title--abstract searches with backward and
forward citation tracking anchored on Ego4D~\citep{grauman2022ego4d},
EPIC-KITCHENS-100~\citep{damen2022epic100},
AntGPT~\citep{zhao2024antgpt}, and
PlausiVL~\citep{mittal2024plausivl}, producing an initial 61-work set.
Window~3 (1 May--31 July 2026) targeted recent challenge reports,
preprints, and lineage records; its 10 evidence records are reported
separately. Two prospective verification passes, completed on 2 and 3
August and restricted to works publicly available by 31 July 2026,
used expanded forecasting, trajectory, adaptation, multimodal-context,
and uncertainty terms plus citation chaining. They added eight
pre-cutoff works, yielding a final retained synthesis set of 69. The
executed verification queries and dispositions are archived in
Supplementary Section~S3 and
\texttt{verification\_search\_log.csv}; earlier-window query templates
are reconstructions from authors' notes and do not reproduce unavailable
initial attrition counts.

The searched sources covered major computer-vision, machine-learning,
AI, and robotics venues; relevant journals; and arXiv cs.CV/cs.AI.
Foundational pre-2018 studies were added through backward citation
tracking. Search sources, reconstructed templates, verification queries,
eligibility rules, and candidate dispositions are reported in
Supplementary Section~S3.

\textbf{Eligibility and screening.}
We include video-based action anticipation studies evaluated under a
recognised anticipation setting, together with third-person procedural
studies and explicitly labelled grounding/planning boundary cases that
clarify the design space. Pure recognition, future-frame prediction, and
robotics/VLA work without direct anticipation evidence or a specific
design implication are excluded. Two authors independently screened
titles and abstracts, resolved disagreements by discussion, and jointly
assessed borderline full texts. The review is therefore presented as a
structured, scoping, design-centric synthesis. It is not presented as a
PRISMA-compliant systematic review, and no exhaustive-coverage claim is
made. Its reporting nevertheless draws on established principles of
transparent evidence synthesis and reproducible machine-learning
reporting. We adapt these principles but do not claim compliance with
formal systematic-review or clinical evidence-grading frameworks
\citep{kitchenham2007slr,guyatt2008grade,pineau2021reproducibility}.

\textbf{Focal selection and coding.}
The focal set contains recurring or influential mechanisms; it is not a
list of every published variant. A core focal method must report a recognised
anticipation protocol and additionally contribute design novelty,
component-level ablation, or documented lineage influence. A small set
of boundary cases is admitted only when it introduces a distinctive
feasibility or executability mechanism. Primary-C1 changes the observed
context representation; primary-C2 infers or supplies a goal, intention,
semantic prototype, or trajectory-based intent proxy; primary-C3 changes
the future objective, scorer, sequence model, or decoding stability; and
primary-C4 evaluates or constrains feasibility beyond ordinary label
prediction. Secondary roles are multi-label. The operational manual,
condition-satisfaction matrix, row-level rationale, and alternate-assignment sensitivity analysis are provided in Supplementary
Sections~S1, S4, and S6.

\textbf{Evidence extraction and limitations.}
Protocol metadata, quantitative rows, publication maturity, component
isolation, protocol relevance, reproduction status, and source locators
were extracted from primary papers where available. Focal quantitative
and ablation entries received a second-pass source check; unresolved
fields were omitted or marked as source-described. The study was not
prospectively designed for duplicate extraction, and the final coding
matrix was not blindly recoded in full by a second author, so no
inter-rater coefficient is claimed. These limitations are mitigated by
row-level release, explicit coding rules, a claim--evidence map, and
sensitivity analysis, but they are not treated as substitutes for an
independent reliability study.

\subsection{Problem Definition and Notation}

For expository consistency, we adopt a unified notation across
the surveyed literature. The formulations below are schematic
abstractions intended to place diverse methods within a common
probabilistic view. They are not verbatim reproductions of any single
paper.

\begin{table}[h]
\scriptsize\centering
\caption{Notation used throughout this survey.}
\label{tab:notation}
\setlength{\tabcolsep}{4pt}
\renewcommand{\arraystretch}{1.15}
\begin{tabular}{cl}
\toprule
\textbf{Symbol} & \textbf{Meaning} \\ \midrule
$X_{1:t}$       & Observed video prefix \\
$\tilde{X}$     & Context-enriched representation \\
$Y_{t+1:T}$     & Future action sequence (verb--noun pairs) \\
$G$             & Latent goal or high-level intention \\
$S$             & Grounding state (program/scene graph/trajectory) \\
$\mathcal{F}$   & Feasible set of futures consistent with $S$ \\
$N, Z, K$       & Context segments, horizon, candidate count \\
ED              & Minimum Edit Distance ($\downarrow$) \\
seq-F1          & Sequence-level F1 ($\uparrow$) \\
MT5R            & Mean Top-5 Recall, class-mean ($\uparrow$) \\
mAP             & Mean Average Precision ($\uparrow$) \\
\bottomrule
\end{tabular}
\end{table}

Let $X_{1:t}$ denote the observed prefix and $Y_{t+1:T}$ the
target future action sequence. A generic anticipation formulation
can be written as
\begin{equation}
  \hat{Y}_{t+1:T}
  = \arg\max_{Y_{t+1:T}} P(Y_{t+1:T}\mid X_{1:t}).
  \label{eq:base}
\end{equation}

A conceptual goal-conditioned extension introduces a latent goal
variable $G$, written here for a discrete goal space; for a
continuous goal the summation is replaced by an expectation:
\begin{equation}
  \hat{Y}_{t+1:T}
  = \arg\max_{Y_{t+1:T}}
  \sum_G P(Y_{t+1:T}\mid X_{1:t}, G)\,P(G\mid X_{1:t}).
  \label{eq:goal}
\end{equation}

In practice, many goal-conditioned methods approximate this
through MAP-style inference, selecting a single best-estimate
goal $\hat{G}$ rather than marginalising over multiple plausible
goals. The gap between full marginalisation and single-goal
approximation is one source of goal-hallucination errors
discussed in Section~\ref{sec:c2}.

Grounded prediction further constrains decoding to a feasible set
$\mathcal{F}(X_{1:t}, S)$ derived from a grounding state $S$:
\begin{equation}
  \hat{Y}_{t+1:T}
  = \arg\max_{Y_{t+1:T}\in \mathcal{F}(X_{1:t},S)}
  P(Y_{t+1:T}\mid X_{1:t}, G, S).
  \label{eq:grounded}
\end{equation}

These abstractions motivate the component-wise view adopted in
the remainder of the survey and provide a common notation for
comparing methods that differ substantially in architecture and
training strategy.

\subsection{Task-Regime Taxonomy Before Model Taxonomy}
\label{sec:task-regimes}

Insertion-point labels are meaningful only after the prediction task is
specified. The reviewed literature combines regimes that differ in
more than horizon, so each study is coded along six operational axes
before assigning C1--C4. Table~\ref{tab:task-regimes} prevents an
EK-100 single-label classifier, an Ego4D-LTA sequence forecaster, an
object-interaction localiser, and an executable planner from being
treated as points on one scalar difficulty axis.

\begin{table*}[t]
\footnotesize\centering
\caption{Operational task-regime taxonomy applied before C1--C4
coding. ``Language-augmented'' denotes an identifiable language- or
VLM-derived signal; it does not imply an autoregressive LLM at
inference.}
\label{tab:task-regimes}
\setlength{\tabcolsep}{4pt}
\renewcommand{\arraystretch}{1.18}
\begin{tabularx}{\textwidth}{p{2.0cm}p{3.0cm}Yp{3.0cm}p{3.5cm}}
\toprule
\textbf{Axis} & \textbf{Operational values} & \textbf{Examples} &
\textbf{Required reporting} & \textbf{Interpretive consequence} \\
\midrule
Observation & Pre-action clip; recognised history; single frame;
partial ongoing action; cross-view stream & EK-100; Ego4D-LTA; AAG;
Ego4D-STA; DCPGN & Window length, FPS, observed labels, view and
adaptation access & Fixes what evidence is available before any
language component acts \\
Target & One label; action sequence; verb+noun+time/object; open-text
step; executable plan & EK-100; Ego4D-LTA/STA; TrajPilot; Anticipate
\& Act & Target schema, sequence length, localisation or solvability
requirement & Changes the output space and meaning of ``correct'' \\
Vocabulary & Closed; open; hybrid/retrieval-constrained & EK-100;
Ego-Exo4D open vocabulary; prototype methods & Class inventory,
unseen/rare-class policy, text-to-label mapping & Separates long-tail
difficulty from genuine open-vocabulary generalisation \\
Horizon & Anticipation time $\tau_a$; discrete horizon $Z$; real-time
rollout & EK-100; Ego4D-LTA; streaming/planning studies & $\tau_a$ or
$Z$, observation length $N$, candidates $K$, sampling policy & Prevents
single-action and sequence protocols from being merged \\
Goal access & Provided; masked; inferred latent goal; semantic
conditioning only; absent & Goal-conditioned planning; TrajPilot goal
masking; AntGPT; PAR-VLA & Whether a goal is observed, inferred, or
approximated by prototypes/text & Distinguishes latent-goal inference
from semantic feature conditioning \\
Evaluation & MT5R; edit distance; mAP; calibration/coverage;
solvability/success & EK-100; Ego4D-LTA/STA; Du et al.; planning
boundary studies & Evaluator, split, version, subset and source locator
& Metrics across task families are not interchangeable \\
\bottomrule
\end{tabularx}
\end{table*}

\begin{table*}[t]
\footnotesize\centering
\caption{Task regime $\times$ insertion-point matrix. Filled cells
summarise common analytical roles, not requirements. C4 is an
adjacent/emerging extension whose outputs may be constraints, plans,
or feasibility judgements rather than benchmark action labels.}
\label{tab:task-by-component}
\setlength{\tabcolsep}{5pt}
\renewcommand{\arraystretch}{1.18}
\begin{tabularx}{\textwidth}{p{3.4cm}YYYY}
\toprule
\textbf{Task family} & \textbf{C1 context} & \textbf{C2 goal/intention} &
\textbf{C3 decoding} & \textbf{C4 adjacent grounding/planning} \\
\midrule
Single-action, closed vocabulary & Backbone, modalities, history &
Optional semantic/intention conditioning & Class scoring & Usually not
required \\
Long-term sequence forecasting & Observed-action/video history &
Latent or prompted goal/subgoal & Autoregressive or structured sequence
prediction & Feasibility/procedural constraints \\
Object-interaction anticipation & Actor/object context, localisation &
Interaction intent or prototypes & Verb/noun/time/box prediction &
Geometric or affordance consistency \\
Cross-view/test-time adaptation & View-normalised context and memory &
Prototype/textual progression clues & Multi-label anticipation &
Optional adaptation constraints \\
Planning/robotic boundary & Perceptual state and history & Explicit goal
or subgoal & Plan/step generation & Solvability, dynamics, execution \\
\bottomrule
\end{tabularx}
\end{table*}

\section{Evolution of Action Anticipation}
\label{sec:evolution}

We organise the development of action anticipation into four
broad phases, each characterised by a dominant unresolved
limitation, not by a rigidly exclusive paradigm. These
limitations align with the four functional components of our
taxonomy: context modelling~(C1), intention modelling~(C2),
future decoding~(C3), and grounding or executability
constraints~(C4). This periodisation is an interpretive historical reading, not an
empirically estimated sequence of bottlenecks. Its purpose is not
to impose a strict boundary, but to provide an interpretive lens for understanding how successive lines of work
shifted the field's centre of difficulty.

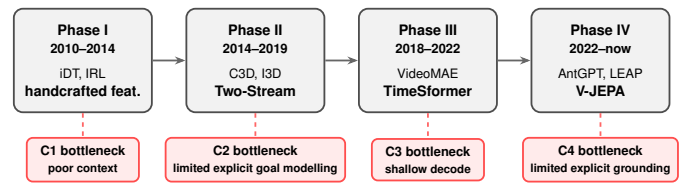
\begin{figure}[t]
\centering
\resizebox{\linewidth}{!}{%
\begin{tikzpicture}[
  font=\sffamily\bfseries,
  phase/.style={rectangle, rounded corners=6pt,
    draw=black!70, very thick, fill=gray!10,
    minimum width=3.2cm, minimum height=2.4cm,
    inner sep=7pt, align=center},
  bn/.style={rectangle, rounded corners=4pt,
    draw=red!70, very thick, fill=red!8,
    minimum width=2.6cm, minimum height=0.8cm,
    inner sep=5pt, align=center, font=\small\sffamily\bfseries},
  arr/.style={-{Stealth[length=7pt]}, very thick, draw=black!60}
]
\node[phase] (p1) at (0,0)
  {\normalsize Phase I\\{\small 2010--2014}\\[4pt]
   \small\mdseries iDT, IRL\\handcrafted feat.};
\node[phase] (p2) at (4,0)
  {\normalsize Phase II\\{\small 2014--2019}\\[4pt]
   \small\mdseries C3D, I3D\\Two-Stream};
\node[phase] (p3) at (8,0)
  {\normalsize Phase III\\{\small 2018--2022}\\[4pt]
   \small\mdseries VideoMAE\\TimeSformer};
\node[phase] (p4) at (12,0)
  {\normalsize Phase IV\\{\small 2022--now}\\[4pt]
   \small\mdseries AntGPT, LEAP\\V-JEPA};
\node[bn] (b1) at (0,-2.3) {\small C1 bottleneck\\\footnotesize poor context};
\node[bn] (b2) at (4,-2.3) {\small C2 bottleneck\\\footnotesize limited explicit goal modelling};
\node[bn] (b3) at (8,-2.3) {\small C3 bottleneck\\\footnotesize shallow decode};
\node[bn] (b4) at (12,-2.3) {\small C4 bottleneck\\\footnotesize limited explicit grounding};
\draw[arr] (p1.east) -- (p2.west);
\draw[arr] (p2.east) -- (p3.west);
\draw[arr] (p3.east) -- (p4.west);
\draw[dashed, very thick, draw=red!60] (p1.south) -- (b1.north);
\draw[dashed, very thick, draw=red!60] (p2.south) -- (b2.north);
\draw[dashed, very thick, draw=red!60] (p3.south) -- (b3.north);
\draw[dashed, very thick, draw=red!60] (p4.south) -- (b4.north);
\end{tikzpicture}
}
\caption{An interpretive four-phase reading of the evolution of action anticipation; it is not an empirically estimated periodisation. Solid arrows indicate broad chronological
progression, while dashed links indicate the dominant unresolved
limitation that remained salient in each phase. The C1--C4 labels
align these historical shifts with the functional components of
the taxonomy in Section~\ref{sec:pipeline}.}
\label{fig:phases}
\end{figure}

\subsection{Phase I: Limited Contextual Representation (2010--2014)}

Early work, including iDT~\citep{wang2013idt}, early recognition
protocols~\citep{ryoo2011early}, and IRL-based activity
prediction~\citep{kitani2012activity}, relied on handcrafted or
weakly semantic representations extracted from relatively short
temporal windows. Although these approaches established the
problem setting, they offered only limited access to broader
context, making context representation the dominant unresolved
difficulty in this phase.

\subsection{Phase II: Stronger Context, Weak Intention Modelling (2014--2019)}

The transition to learned spatiotemporal representations,
including Two-Stream CNNs~\citep{simonyan2014twostream},
C3D~\citep{tran2015c3d}, and I3D~\citep{carreira2017quovadis},
substantially improved the quality of contextual encoding and
helped relax the earlier C1 limitation. Building on this shift,
dedicated anticipation models began to emerge, including the
temporal occurrence framework of Abu~Farha
et~al.~\citep{abufarha2018whenwillyou} and the rolling--unrolling
LSTM formulation of Furnari and
Farinella~\citep{furnari2019rulstm} for EPIC-KITCHENS. However,
these methods still modelled future actions primarily through
observed temporal patterns, with limited explicit treatment of
high-level intention or goal structure. In this sense, intention
modelling remained the dominant unresolved limitation.

\subsection{Phase III: Richer Encoders, Still Shallow Future Decoding (2018--2022)}

Transformer-based and self-supervised video representations,
including TimeSformer~\citep{bertasius2021timesformer},
ViViT~\citep{arnab2021vivit}, and
VideoMAE~\citep{tong2022videomae} and VideoMAE~V2~\citep{wang2023videomaev2}, expanded the capacity of
anticipation systems to encode longer-range and semantically
richer context. In parallel, benchmarks such as
EPIC-KITCHENS~\citep{damen2022epic100} and
Ego4D~\citep{grauman2022ego4d} helped stabilise evaluation and
made long-horizon anticipation a more explicit target. Yet in
many methods, future prediction still relied on relatively
shallow decoding of action labels from enriched visual context,
rather than on structured generation over plausible procedural
futures. As a result, future decoding remained the dominant
unresolved limitation in this phase.

\subsection{Phase IV: Goal-Conditioned and Grounded Prediction (2022--present)}

Recent work has increasingly targeted long-horizon anticipation
through goal-aware, language-augmented, or plausibility-aware
prediction. Methods such as AntGPT~\citep{zhao2024antgpt} and
PlausiVL~\citep{mittal2024plausivl} illustrate a shift from
purely correlational next-action prediction toward more explicit
reasoning about likely future trajectories. At the same time,
this shift has made a further limitation more visible: predictions
may be fluent and goal-consistent while still violating domain
constraints or executability requirements. This motivates the
C4 perspective, under which grounding and executability become
central rather than auxiliary concerns.

At the same time, recent dense-pretraining work in the V-JEPA line
exposes a comparability problem that is central to this review:
published short-horizon values cannot be separated cleanly from
pretraining scale, probe design, loss, sampling policy, and evaluator
implementation. Because the relevant reports are recent and partly
preprint-based, we treat them as an emerging reporting incompatibility rather
than as a settled comparison between language and non-language model
families. The recent literature does not identify a single component as
the unique determinant of performance. Instead, results depend on
interactions among context, intention, decoding, and grounding.

\subsection{Historical Synthesis}

What this four-phase reading ultimately reveals is that the
field's centre of difficulty has shifted without being resolved. Each generation of methods removed one dominant
bottleneck only to expose the next: stronger contextual encoders
made the absence of goal reasoning visible; goal-conditioned
prediction made shallow decoding visible; and fluent,
goal-consistent generation has now made the absence of grounding
and executability constraints visible. A plausible interpretation is
that limited procedural supervision constrained many pre-LLM systems;
architectural inexpressivity alone does not explain the pattern. LLM-era
methods provide one route to broader priors, while simultaneously
changing model scale, pretraining, optimisation, and multimodal
interfaces; the historical pattern should therefore be read as an
interpretive synthesis, not a controlled causal claim. The C1--C4 taxonomy developed in the next section makes
this shifting-bottleneck pattern explicit and provides the frame
within which the remainder of the survey is organised.

\section{The Generic Pipeline}
\label{sec:pipeline}

\colorlet{hC1dark}{teal!70!black}
\colorlet{hC1mid}{teal!52}
\colorlet{hC1lite}{teal!22}
\colorlet{hC2dark}{violet!70!black}
\colorlet{hC2mid}{violet!52}
\colorlet{hC2lite}{violet!22}
\colorlet{hC3dark}{orange!88!black}
\colorlet{hC3mid}{orange!65}
\colorlet{hC3lite}{orange!28}
\colorlet{hC4dark}{red!72!black}
\colorlet{hC4mid}{red!55}
\colorlet{hC4lite}{red!24}

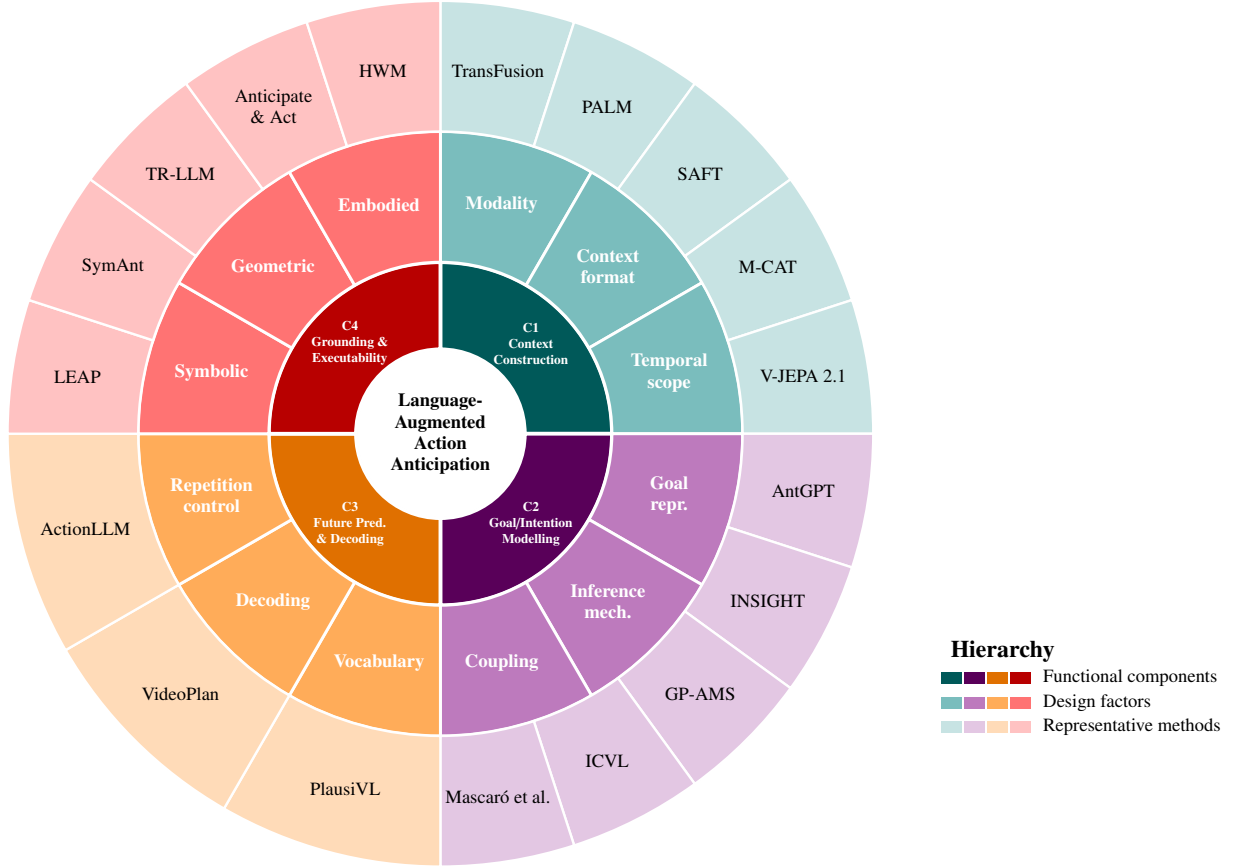
\begin{figure*}[t]
\centering
\begin{tikzpicture}[font=\sffamily, scale=0.72]

  \fill[white] (0,0) circle (1.55);
  \draw[black!18, line width=0.8pt] (0,0) circle (1.55);
  \node[align=center, font=\bfseries\scriptsize] at (0,0)
    {Language-\\Augmented\\Action\\Anticipation};
  \node[align=center, font=\tiny, text=black!65] at (0,-1.05){};
  
  \path[fill=hC1dark, draw=white, line width=1.4pt]
    (90:1.55) arc(90:0:1.55) -- (0:3.15) arc(0:90:3.15) -- cycle;
  \node[align=center, text=white, font=\bfseries\tiny]
    at (45:2.35) {\text{C1}\\Context\\Construction};
  \path[fill=hC2dark, draw=white, line width=1.4pt]
    (0:1.55) arc(0:-90:1.55) -- (-90:3.15) arc(-90:0:3.15) -- cycle;
  \node[align=center, text=white, font=\bfseries\tiny]
    at (-45:2.35) {\text{C2}\\Goal/Intention\\Modelling};
  \path[fill=hC3dark, draw=white, line width=1.4pt]
    (-90:1.55) arc(-90:-180:1.55) -- (-180:3.15) arc(-180:-90:3.15) -- cycle;
  \node[align=center, text=white, font=\bfseries\tiny]
    at (-135:2.35) {\text{C3}\\Future Pred.\\\& Decoding};
  \path[fill=hC4dark, draw=white, line width=1.4pt]
    (180:1.55) arc(180:90:1.55) -- (90:3.15) arc(90:180:3.15) -- cycle;
  \node[align=center, text=white, font=\bfseries\tiny]
    at (135:2.35) {C4\\Grounding \&\\Executability};

  \path[fill=hC1mid, draw=white, line width=1.1pt]
    (90:3.15) arc(90:60:3.15) -- (60:5.55) arc(60:90:5.55) -- cycle;
  \node[align=center, text=white, font=\bfseries\scriptsize] at (75:4.35) {Modality};
  \path[fill=hC1mid, draw=white, line width=1.1pt]
    (60:3.15) arc(60:30:3.15) -- (30:5.55) arc(30:60:5.55) -- cycle;
  \node[align=center, text=white, font=\bfseries\scriptsize] at (45:4.35) {Context\\format};
  \path[fill=hC1mid, draw=white, line width=1.1pt]
    (30:3.15) arc(30:0:3.15) -- (0:5.55) arc(0:30:5.55) -- cycle;
  \node[align=center, text=white, font=\bfseries\scriptsize] at (15:4.35) {Temporal\\scope};

  \path[fill=hC2mid, draw=white, line width=1.1pt]
    (0:3.15) arc(0:-30:3.15) -- (-30:5.55) arc(-30:0:5.55) -- cycle;
  \node[align=center, text=white, font=\bfseries\scriptsize] at (-15:4.35) {Goal\\repr.};
  \path[fill=hC2mid, draw=white, line width=1.1pt]
    (-30:3.15) arc(-30:-60:3.15) -- (-60:5.55) arc(-60:-30:5.55) -- cycle;
  \node[align=center, text=white, font=\bfseries\scriptsize] at (-45:4.35) {Inference\\mech.};
  \path[fill=hC2mid, draw=white, line width=1.1pt]
    (-60:3.15) arc(-60:-90:3.15) -- (-90:5.55) arc(-90:-60:5.55) -- cycle;
  \node[align=center, text=white, font=\bfseries\scriptsize] at (-75:4.35) {Coupling};

  \path[fill=hC3mid, draw=white, line width=1.1pt]
    (-90:3.15) arc(-90:-120:3.15) -- (-120:5.55) arc(-120:-90:5.55) -- cycle;
  \node[align=center, text=white, font=\bfseries\scriptsize] at (-105:4.35) {Vocabulary};
  \path[fill=hC3mid, draw=white, line width=1.1pt]
    (-120:3.15) arc(-120:-150:3.15) -- (-150:5.55) arc(-150:-120:5.55) -- cycle;
  \node[align=center, text=white, font=\bfseries\scriptsize] at (-135:4.35) {Decoding};
  \path[fill=hC3mid, draw=white, line width=1.1pt]
    (-150:3.15) arc(-150:-180:3.15) -- (-180:5.55) arc(-180:-150:5.55) -- cycle;
  \node[align=center, text=white, font=\bfseries\scriptsize] at (-165:4.35) {Repetition\\control};

  \path[fill=hC4mid, draw=white, line width=1.1pt]
    (180:3.15) arc(180:150:3.15) -- (150:5.55) arc(150:180:5.55) -- cycle;
  \node[align=center, text=white, font=\bfseries\scriptsize] at (165:4.35) {Symbolic};
  \path[fill=hC4mid, draw=white, line width=1.1pt]
    (150:3.15) arc(150:120:3.15) -- (120:5.55) arc(120:150:5.55) -- cycle;
  \node[align=center, text=white, font=\bfseries\scriptsize] at (135:4.35) {Geometric};
  \path[fill=hC4mid, draw=white, line width=1.1pt]
    (120:3.15) arc(120:90:3.15) -- (90:5.55) arc(90:120:5.55) -- cycle;
  \node[align=center, text=white, font=\bfseries\scriptsize] at (105:4.35) {Embodied};

  \path[fill=hC1lite, draw=white, line width=0.9pt]
    (90:5.55) arc(90:72:5.55) -- (72:7.95) arc(72:90:7.95) -- cycle;
  \node[align=center, font=\scriptsize] at (81:6.75) {TransFusion};
  \path[fill=hC1lite, draw=white, line width=0.9pt]
    (72:5.55) arc(72:54:5.55) -- (54:7.95) arc(54:72:7.95) -- cycle;
  \node[align=center, font=\scriptsize] at (63:6.75) {PALM};
  \path[fill=hC1lite, draw=white, line width=0.9pt]
    (54:5.55) arc(54:36:5.55) -- (36:7.95) arc(36:54:7.95) -- cycle;
  \node[align=center, font=\scriptsize] at (45:6.75) {SAFT};
  \path[fill=hC1lite, draw=white, line width=0.9pt]
    (36:5.55) arc(36:18:5.55) -- (18:7.95) arc(18:36:7.95) -- cycle;
  \node[align=center, font=\scriptsize] at (27:6.75) {M-CAT};
  \path[fill=hC1lite, draw=white, line width=0.9pt]
    (18:5.55) arc(18:0:5.55) -- (0:7.95) arc(0:18:7.95) -- cycle;
  \node[align=center, font=\scriptsize] at (9:6.75) {V-JEPA~2.1};

  \path[fill=hC2lite, draw=white, line width=0.9pt]
    (0:5.55) arc(0:-18:5.55) -- (-18:7.95) arc(-18:0:7.95) -- cycle;
  \node[align=center, font=\scriptsize] at (-9:6.75) {AntGPT};
  \path[fill=hC2lite, draw=white, line width=0.9pt]
    (-18:5.55) arc(-18:-36:5.55) -- (-36:7.95) arc(-36:-18:7.95) -- cycle;
  \node[align=center, font=\scriptsize] at (-27:6.75) {INSIGHT};
  \path[fill=hC2lite, draw=white, line width=0.9pt]
    (-36:5.55) arc(-36:-54:5.55) -- (-54:7.95) arc(-54:-36:7.95) -- cycle;
  \node[align=center, font=\scriptsize] at (-45:6.75) {GP-AMS};
  \path[fill=hC2lite, draw=white, line width=0.9pt]
    (-54:5.55) arc(-54:-72:5.55) -- (-72:7.95) arc(-72:-54:7.95) -- cycle;
  \node[align=center, font=\scriptsize] at (-63:6.75) {ICVL};
  \path[fill=hC2lite, draw=white, line width=0.9pt]
    (-72:5.55) arc(-72:-90:5.55) -- (-90:7.95) arc(-90:-72:7.95) -- cycle;
  \node[align=center, font=\scriptsize] at (-81:6.75) {Mascar\'o et~al.};

  \path[fill=hC3lite, draw=white, line width=0.9pt]
    (-90:5.55) arc(-90:-120:5.55) -- (-120:7.95) arc(-120:-90:7.95) -- cycle;
  \node[align=center, font=\scriptsize] at (-105:6.75) {PlausiVL};
  \path[fill=hC3lite, draw=white, line width=0.9pt]
    (-120:5.55) arc(-120:-150:5.55) -- (-150:7.95) arc(-150:-120:7.95) -- cycle;
  \node[align=center, font=\scriptsize] at (-135:6.75) {VideoPlan};
  \path[fill=hC3lite, draw=white, line width=0.9pt]
    (-150:5.55) arc(-150:-180:5.55) -- (-180:7.95) arc(-180:-150:7.95) -- cycle;
  \node[align=center, font=\scriptsize] at (-165:6.75) {ActionLLM};

  \path[fill=hC4lite, draw=white, line width=0.9pt]
    (180:5.55) arc(180:162:5.55) -- (162:7.95) arc(162:180:7.95) -- cycle;
  \node[align=center, font=\scriptsize] at (171:6.75) {LEAP};
  \path[fill=hC4lite, draw=white, line width=0.9pt]
    (162:5.55) arc(162:144:5.55) -- (144:7.95) arc(144:162:7.95) -- cycle;
  \node[align=center, font=\scriptsize] at (153:6.75) {SymAnt};
  \path[fill=hC4lite, draw=white, line width=0.9pt]
    (144:5.55) arc(144:126:5.55) -- (126:7.95) arc(126:144:7.95) -- cycle;
  \node[align=center, font=\scriptsize] at (135:6.75) {TR-LLM};
  \path[fill=hC4lite, draw=white, line width=0.9pt]
    (126:5.55) arc(126:108:5.55) -- (108:7.95) arc(108:126:7.95) -- cycle;
  \node[align=center, font=\scriptsize] at (117:6.75) {Anticipate\\\& Act};
  \path[fill=hC4lite, draw=white, line width=0.9pt]
    (108:5.55) arc(108:90:5.55) -- (90:7.95) arc(90:108:7.95) -- cycle;
  \node[align=center, font=\scriptsize] at (99:6.75) {HWM};

  \begin{scope}[shift={(9.2,-5.8)}]
    \node[anchor=west, font=\bfseries\small] at (0,1.8) {Hierarchy};
    \fill[hC1dark] (0,1.2) rectangle +(0.38,0.22);
    \fill[hC2dark] (0.42,1.2) rectangle +(0.38,0.22);
    \fill[hC3dark] (0.84,1.2) rectangle +(0.38,0.22);
    \fill[hC4dark] (1.26,1.2) rectangle +(0.38,0.22);
    \node[anchor=west, font=\scriptsize] at (1.7,1.31) {Functional components};
    \fill[hC1mid] (0,0.75) rectangle +(0.38,0.22);
    \fill[hC2mid] (0.42,0.75) rectangle +(0.38,0.22);
    \fill[hC3mid] (0.84,0.75) rectangle +(0.38,0.22);
    \fill[hC4mid] (1.26,0.75) rectangle +(0.38,0.22);
    \node[anchor=west, font=\scriptsize] at (1.7,0.86) {Design factors};
    \fill[hC1lite] (0,0.30) rectangle +(0.38,0.22);
    \fill[hC2lite] (0.42,0.30) rectangle +(0.38,0.22);
    \fill[hC3lite] (0.84,0.30) rectangle +(0.38,0.22);
    \fill[hC4lite] (1.26,0.30) rectangle +(0.38,0.22);
    \node[anchor=west, font=\scriptsize] at (1.7,0.41) {Representative methods};
  \end{scope}

\end{tikzpicture}
\caption{Hierarchical design-centric overview of language-augmented action
anticipation.  The centre identifies the task.  \textbf{Ring~1}
(solid colours) shows the four functional components of the C1--C4
taxonomy: context construction (C1, teal), goal/intention modelling
(C2, violet), future prediction and decoding (C3, orange), and
an emerging grounding and executability extension (C4, red).  \textbf{Ring~2} (mid
tones) enumerates the major design factors for each component
(Tables~\ref{tab:c1-factors}--\ref{tab:c4-factors}).  \textbf{Ring~3}
(light tones) lists \emph{selected illustrative} methods; it is not
the complete focal set. The authoritative list of all 25 focal
methods is Table~\ref{tab:taxonomy-overview}.}
\label{fig:hierarchical-overview}
\end{figure*}

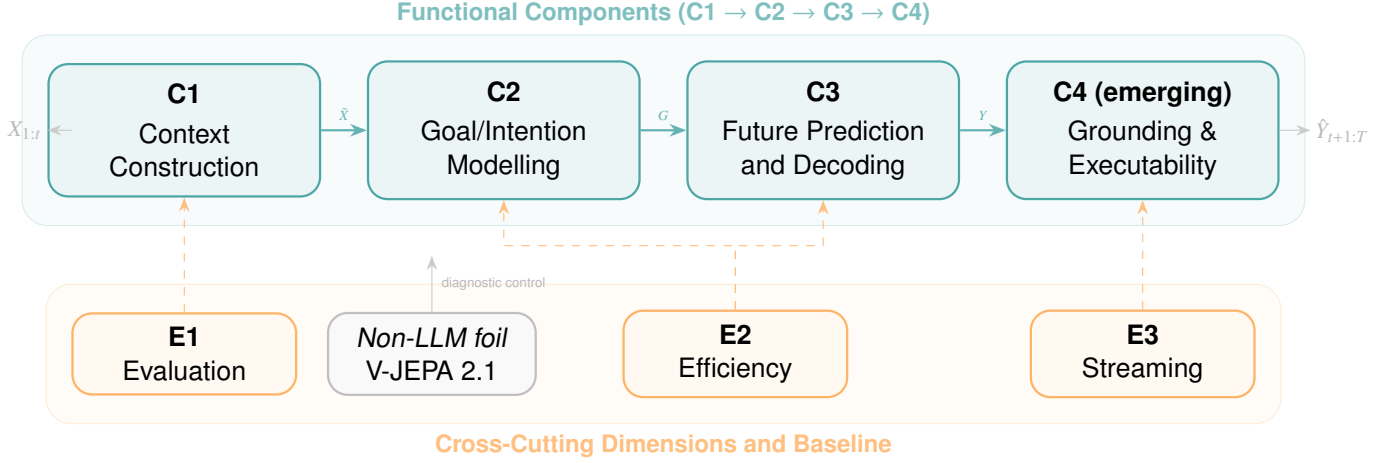
\begin{figure*}[t]
\centering
\resizebox{\linewidth}{!}{%
\begin{tikzpicture}[
  font=\sffamily,
  comp/.style={rectangle, rounded corners=7pt,
    draw=teal!70, thick, fill=teal!8,
    minimum width=3.4cm, minimum height=1.3cm,
    inner sep=7pt, align=center},
  dep/.style={rectangle, rounded corners=7pt,
    draw=orange!60, thick, fill=orange!5,
    minimum width=2.8cm, minimum height=0.95cm,
    inner sep=5pt, align=center},
  foil/.style={rectangle, rounded corners=7pt,
    draw=gray!50, thick, fill=gray!5,
    minimum width=2.6cm, minimum height=0.9cm,
    inner sep=5pt, align=center},
  arr/.style={-{Stealth[length=6pt]}, thick, draw=teal!60},
  darrL/.style={-{Stealth[length=5pt]}, dashed, draw=orange!50},
  garr/.style={-{Stealth[length=5pt]}, draw=gray!40}
]

\node[comp] (C1) at (0,0)
  {\textbf{C1}\\[2pt]Context\\Construction};
\node[comp] (C2) at (4,0)
  {\textbf{C2}\\[2pt]Goal/Intention\\Modelling};
\node[comp] (C3) at (8,0)
  {\textbf{C3}\\[2pt]Future Prediction\\and Decoding};
\node[comp] (C4) at (12,0)
  {\textbf{C4 (emerging)}\\[2pt]Grounding \&\\Executability};

\draw[arr] (C1.east) -- (C2.west)
  node[midway,above,font=\tiny\sffamily,text=teal!60]{$\tilde{X}$};
\draw[arr] (C2.east) -- (C3.west)
  node[midway,above,font=\tiny\sffamily,text=teal!60]{$G$};
\draw[arr] (C3.east) -- (C4.west)
  node[midway,above,font=\tiny\sffamily,text=teal!60]{$Y$};

\node[font=\footnotesize\sffamily,text=gray!55] at (-2,0) {$X_{1:t}$};
\draw[garr] (-1.4,0) -- (C1.west);

\node[font=\footnotesize\sffamily,text=gray!55] at (14.5,0)
  {$\hat{Y}_{t+1:T}$};
\draw[garr] (C4.east) -- (14.1,0);

\node[dep]  (E1)   at (0,-2.8)   {\textbf{E1}\\Evaluation};
\node[foil] (foil) at (3.1,-2.8) {\textit{Non-LLM foil}\\V-JEPA~2.1};
\node[dep]  (E2)   at (6.9,-2.8) {\textbf{E2}\\Efficiency};
\node[dep]  (E3)   at (12,-2.8)  {\textbf{E3}\\Streaming};

\draw[darrL] (E1.north) -- (C1.south);
\draw[darrL] (E2.north) -- ++(0,0.8) -| (C2.south);
\draw[darrL] (E2.north) -- ++(0,0.8) -| (C3.south);
\draw[darrL] (E3.north) -- (C4.south);

\draw[garr] (foil.north) -- ++(0,0.7)
  node[midway,right,font=\tiny\sffamily,text=gray!55]{diagnostic control};

\begin{scope}[on background layer]
  \node[fill=teal!3, draw=teal!15, rounded corners=9pt,
    fit=(C1)(C2)(C3)(C4), inner sep=9pt,
    label={[font=\small\bfseries\sffamily,text=teal!50]above:
      Functional Components (C1 $\to$ C2 $\to$ C3 $\to$ C4)}] {};

  \node[fill=orange!3, draw=orange!15, rounded corners=9pt,
    fit=(E1)(E2)(E3)(foil), inner sep=9pt,
    label={[font=\small\bfseries\sffamily,text=orange!50]below:
      Cross-Cutting Dimensions and Baseline}] {};
\end{scope}

\end{tikzpicture}
}
\caption{Generic pipeline of action anticipation.
\textbf{Top:} four functional components (C1--C4): context
construction produces $\tilde{X}$; goal/intention modelling
infers $G$; future prediction and decoding generates $Y$;
grounding and executability restrict decoding to the feasible set
$\mathcal{F}$, yielding $\hat{Y}_{t+1:T}$.
\textbf{Bottom:} E1 denotes evaluation and protocol comparability,
discussed in Section~\ref{sec:benchmarks}; E2 and E3 denote
deployment-oriented efficiency and streaming constraints, detailed
in Section~\ref{sec:deploy}. The V-JEPA family is included as an
illustrative non-LLM control and as a case study in unresolved
short-horizon reporting comparability; its published values do not
establish a shared-code ranking against language-augmented systems.}
\label{fig:pipeline}
\end{figure*}

Reviewed methods can be organised within a generic pipeline
comprising four functional components and three cross-cutting
dimensions (Fig.~\ref{fig:pipeline}). This decomposition is used
as an analytical framework for comparison. It does not assume that all
methods instantiate fully separable modules.
Figure~\ref{fig:hierarchical-overview} provides a complementary
hierarchical view of the same structure, mapping each functional
component to its major design factors and representative methods.

The four-component structure is compatible with the probabilistic
formulation in Section~\ref{sec:scope}: C1 represents the construction
of $\tilde{X}$ from $X_{1:t}$; C2 represents inference over $G$; C3
represents decoding of $Y$; and C4 represents restriction or checking
through $\mathcal{F}$. This correspondence motivates the taxonomy but
does not prove uniqueness. Alternative decompositions could separate
perception from temporal aggregation, goal inference from uncertainty
modelling, or scene consistency from physical executability. We use
four components because they provide a compact, explicit, and auditable coding
frame for the surveyed mechanisms, not because five- or six-component
alternatives are mathematically invalid. Supplementary Section~S6
reports how plausible alternate primary assignments affect the
component counts.

A generic context--reasoning--prediction--planning pipeline describes
information flow. The C1--C4 map is instead an attribution and diagnosis
framework. Each component specifies (i) the intervention whose marginal
effect should be tested, (ii) the counterfactual control needed to test
it, (iii) the characteristic failure signature, and (iv) the evidence
claim that remains defensible when other components are unmatched. Thus
a C1 claim requires a representation-matched decoder comparison, a C2
claim requires an explicit goal/intention ablation, a C3 claim requires
matched context and goal inputs, and a C4 claim requires a feasibility-sensitive outcome in addition to label agreement.

\begin{table*}[t]
\scriptsize
\centering
\caption{Central evidence-aware design map linking each component to its intervention, required counterfactual, characteristic failure signature, and permissible evidence claim.}
\label{tab:central-design-map}
\setlength{\tabcolsep}{4.6pt}
\renewcommand{\arraystretch}{1.24}
\begin{tabularx}{\textwidth}{P{0.75cm}P{2.75cm}P{3.65cm}P{3.05cm}Y}
\toprule
\textbf{Comp.} & \textbf{Intervention} & \shortstack[l]{\textbf{Required}\\\textbf{counterfactual}} & \shortstack[l]{\textbf{Characteristic}\\\textbf{failure signature}} & \shortstack[l]{\textbf{Permissible}\\\textbf{evidence claim}} \\
\midrule
C1 & Context or representation change & Hold the decoder and goal pathway fixed; match training data and evaluation code where possible & Context noise, missing evidence, modality mismatch, or weak temporal representation & Representation-level contribution under the stated matched controls \\
C2 & Goal, intention, or semantic-conditioning signal & Hold C1 and C3 fixed; remove, replace, or perturb the goal/intention signal & Goal hallucination, overconstraint, or collapse to an incorrect procedural branch & Goal-conditioning contribution, not a general LLM effect \\
C3 & Objective, dependency model, or decoder change & Hold context and goal inputs fixed; match candidate space and decoding budget & Repetition, exposure drift, weak action dependencies, or poor calibration & Decoding-stability or sequence-modelling contribution \\
C4 & Adjacent feasibility, grounding, or planning constraint & Use the same candidate generator without the check, constraint, or planner & Invalid, inconsistent, unsafe, or unreachable futures & Feasibility-sensitive contribution on an outcome that measures executability \\
\bottomrule
\end{tabularx}
\vspace{-2mm}
\end{table*}

This framing yields three insights that a generic pipeline alone does
not. First, it separates model family from functional role: offline
language-derived prototypes, runtime LLM prompting, and non-LLM semantic
reasoning can be compared at the same intervention point. Second, it
shows that the likely bottleneck and the necessary control change with task regime; horizon alone is insufficient. Third, it exposes
cross-component failure propagation and explains why a benchmark number
may support a system-level claim while remaining insufficient for a
component-level attribution. The taxonomy is therefore useful as a map
of comparisons and falsifiable questions, not merely as a list of
modules. Supplementary Section~S6 tests the sensitivity of the high-level
findings to plausible alternate primary assignments.

The two deployment dimensions, E2 (efficiency) and E3 (streaming),
are previewed in Fig.~\ref{fig:pipeline} as cross-cutting concerns
and receive full treatment in Section~\ref{sec:deploy}. Readers
designing systems for edge devices, wearable platforms, or robotic
assistants should note that E2 and E3 interact strongly with the
C1 and C4 choices analysed in the coming sections.

Table~\ref{tab:taxonomy-overview} maps the 25 focal methods,
diagnostic exemplars, and boundary cases analysed in depth across the
four components. Together with the deployment-oriented methods in
Table~\ref{tab:e23}, these tables cover the focal and
deployment-oriented method sets discussed in depth; Supplementary
Table~S1 provides the broader inventory, including lineage and
contextual records discussed elsewhere in the main text.
V-JEPA~2.1~\citep{murLabadia2026vjepa21} is retained as the focal
non-LLM C1 foil, while V-JEPA~2~\citep{assran2025vjepa2} is
discussed as contextual and benchmark evidence. Both are preprints,
and their numbers should be treated as emerging evidence.

Focal-group membership is orthogonal to primary component assignment.
The six non-LLM diagnostic/exemplar methods are distributed across the
same functional map: V-JEPA~2.1 and EgoAnticipator are C1 controls;
Mascar\'o et al. and TrajPilot are C2 controls; AGA is a C3 control;
and SymAnt is a C4 neuro-symbolic exemplar. The three boundary cases
are separately identified because they transfer grounding or planning
mechanisms without satisfying the core anticipation eligibility rule.

\begin{table*}[t]
\scriptsize\centering
\caption{Taxonomy of 25 focal entries: 16 core LLM/VLM-augmented
methods, six non-LLM diagnostic/exemplar methods, and three
planning/robotics boundary cases, distributed across four functional components.  Each method is listed under its primary
component (largest reported marginal gain per the authors' own
ablation, or the component most directly addressed by the stated
contribution).  $\bullet$~=~primary; $\circ$~=~secondary role.
V-JEPA~2.1 is classified under C1 because its primary contribution
is video self-supervised pretraining and it reports directly on
EK-100 at $\tau_a{=}1$s under action MT5R~\citep{murLabadia2026vjepa21}.
InternVideo2~\citep{wang2024internvideo2} is classified as a
non-focal archival baseline because it reports on EK-100 under
Action@1s rather than class-mean MT5R, a different metric not
directly comparable to the V-JEPA rows.
$^\dagger$SymAnt is an unpublished Master's thesis / technical
report rather than a peer-reviewed venue publication
(Section~\ref{sec:c4}). It is retained as a non-LLM neuro-symbolic
C4 diagnostic exemplar: the evidence inventory establishes its
scene- and knowledge-graph mechanism, but not an identifiable
LLM/VLM component. Its inclusion is justified by design relevance,
not by archival evidential weight, and its publication status should be read as
non-archival throughout.
C4 combines benchmark-grounded studies with planning-oriented
boundary evidence and is therefore less mature than C1--C3; publication
status and evidence dimensions are reported in Supplementary Tables~S1 and
S4. Deployment-oriented methods (E2/E3) appear in
Table~\ref{tab:e23}.}
\label{tab:taxonomy-overview}
\setlength{\tabcolsep}{5.0pt}
\renewcommand{\arraystretch}{1.25}
\begin{tabularx}{\textwidth}{P{2.65cm}Ycccc}
\toprule
\textbf{Method} & \textbf{Core idea} & \textbf{C1} & \textbf{C2}
  & \textbf{C3} & \textbf{C4} \\ \midrule
\multicolumn{6}{l}{\textit{Context construction (C1)}} \\
TransFusion~\citep{pasca2024transfusion}
  & Language summaries of past context   & $\bullet$ & $\circ$ & --- & --- \\
PALM~\citep{kim2024palm}
  & MMR-based context + prompt design    & $\bullet$ & --- & $\circ$ & --- \\
SAFT~\citep{saft2026}
  & Sequential textual correction memory & $\bullet$ & --- & $\circ$ & --- \\
M-CAT~\citep{beedu2024mcat}
  & Dual-role text (past+future)         & $\bullet$ & --- & $\circ$ & --- \\
AAG~\citep{benavent2026aag}
  & Single-frame RGB/depth + action-history semantics & $\bullet$ & $\circ$ & --- & --- \\
DCPGN~\citep{shi2026dcpg}
  & Test-time ego--exo adaptation with textual/visual clues & $\bullet$ & $\circ$ & --- & --- \\
V-JEPA~2.1~\citep{murLabadia2026vjepa21}
  & Dense pretraining (non-LLM foil)     & $\bullet$ & --- & --- & --- \\
EgoAnticipator~\citep{chen2025egoanticipator}
  & Retentive/predictive pretraining (non-LLM) & $\bullet$ & --- & $\circ$ & --- \\
\midrule
\multicolumn{6}{l}{\textit{Goal and intention modelling (C2)}} \\
AntGPT~\citep{zhao2024antgpt}
  & Top-down goal inference              & $\circ$ & $\bullet$ & $\circ$ & --- \\
INSIGHT~\citep{insight2026}
  & RL cognitive reasoning chain         & --- & $\bullet$ & --- & --- \\
GP-AMS~\citep{gpams2026}
  & Goal-guided prompting                & --- & $\bullet$ & --- & --- \\
ICVL~\citep{cao2025icvl}
  & VLM intention + LLM decoding         & --- & $\bullet$ & $\circ$ & --- \\
PAR-VLA~\citep{shao2026parvla}
  & Disentangled verb/noun VL prototypes & $\circ$ & $\bullet$ & $\circ$ & --- \\
Mascaró et al.~\citep{mascaro2023intention}
  & Intention-conditioned VAE            & --- & $\bullet$ & --- & --- \\
TrajPilot~\citep{jun2026trajpilot}
  & Predicted camera trajectory as intent-conditioning signal & --- & $\bullet$ & $\circ$ & $\circ$ \\
\midrule
\multicolumn{6}{l}{\textit{Future prediction and decoding (C3)}} \\
PlausiVL~\citep{mittal2024plausivl}
  & Plausibility + anti-repetition loss  & --- & --- & $\bullet$ & $\circ$ \\
VideoPlan~\citep{videoplan2026}
  & Auxiliary tasks + multi-token pred.  & --- & $\circ$ & $\bullet$ & --- \\
ActionLLM~\citep{wu2025actionllm}
  & Cross-modality interaction block     & $\circ$ & --- & $\bullet$ & --- \\
AGA~\citep{tai2026aga}
  & Action-guided attention (non-LLM)    & --- & $\circ$ & $\bullet$ & --- \\
\midrule
\multicolumn{6}{l}{\textit{Emerging grounding and executability extension (C4)}} \\
LEAP~\citep{dessalene2023leap}
  & Executable action programs           & $\circ$ & --- & --- & $\bullet$ \\
FactCheck~\citep{cao2026factcheck}
  & Observe--Plan--Verify agent loop     & $\circ$ & $\circ$ & --- & $\bullet$ \\
SymAnt~\citep{aryan2025symant}$^\dagger$
  & Scene+KG graph search                & --- & --- & --- & $\bullet$ \\
TR-LLM~\citep{trllm2025}
  & Trajectory+LLM fusion                & --- & --- & --- & $\bullet$ \\
Anticipate \& Act~\citep{arora2025anticipate}
  & PDDL joint planning                  & --- & $\circ$ & --- & $\bullet$ \\
HWM~\citep{zhang2026hwm}
  & Hierarchical latent MPC              & --- & --- & --- & $\bullet$ \\
\bottomrule
\end{tabularx}
\end{table*}

\begin{table*}[t]
\footnotesize\centering
\caption{Deployment-oriented methods (E2: efficiency; E3: streaming),
reviewed in Section~\ref{sec:deploy}. These cross-cutting methods are
not assigned to C1--C4. ``Direct'' denotes an anticipation evaluation;
``transfer'' denotes a mechanism imported from efficient or streaming
VideoLLM research without direct anticipation-performance evidence in
the retained source.}
\label{tab:e23}
\setlength{\tabcolsep}{3.5pt}
\renewcommand{\arraystretch}{1.18}
\begin{tabularx}{\textwidth}{lYccp{3.4cm}}
\toprule
\textbf{Method} & \textbf{Core idea} & \textbf{E2} & \textbf{E3}
  & \textbf{Relation to anticipation} \\ \midrule
SparseVLM~\citep{sparsevlm}
  & Text-aware visual token pruning   & $\bullet$ & --- & Transfer mechanism only \\
PruneVid~\citep{prunevid}
  & Static/dynamic token disentanglement & $\bullet$ & --- & Transfer mechanism only \\
VTS~\citep{vts}
  & Key-frame saliency+novelty pruning & $\bullet$ & --- & Transfer mechanism only \\
AdaCM$^2$~\citep{adacm2}
  & Goal-aware rolling memory compression & $\bullet$ & $\circ$ & Transfer mechanism only \\
ReKV~\citep{di2025rekv}
  & Sliding-window KV-cache retrieval  & $\bullet$ & $\bullet$ & Transfer mechanism only \\
CLAM~\citep{zhong2026clam}
  & Cross linear attentive memory      & $\bullet$ & $\circ$ & Direct anticipation evaluation \\
VideoLLM-online~\citep{chen2024videollm}
  & Streaming-EOS silent-by-default    & --- & $\bullet$ & Transfer mechanism only \\
STREAMMIND~\citep{ding2025streammind}
  & Event-gated LLM cognition (source-reported up to 100 FPS) & --- & $\bullet$ & Transfer mechanism only \\
MA-LMM~\citep{he2024malmm}
  & Compressed long-term memory bank  & $\circ$ & $\bullet$ & Transfer mechanism only \\
ProVideLLM~\citep{chatterjee2025provide}
  & Interleaved text+visual cache     & $\circ$ & $\bullet$ & Transfer mechanism only \\
\bottomrule
\end{tabularx}
\end{table*}

\begin{figure*}[t]
\centering
\resizebox{\textwidth}{!}{%
\begin{tikzpicture}[font=\sffamily\scriptsize]
\begin{scope}[shift={(0,0)}]
  \node[anchor=west,font=\bfseries\small] at (0,8.8) {(a) Primary component counts};
  \draw[->,black!55] (0,0) -- (0,8.3);
  \draw[->,black!55] (0,0) -- (5.2,0);
  \foreach \x/\h/\lab in {0.7/8/C1,1.8/7/C2,2.9/4/C3,4.0/6/C4}{
    \fill[teal!35] (\x,0) rectangle +(0.65,\h);
    \draw[teal!70!black] (\x,0) rectangle +(0.65,\h);
    \node at (\x+0.325,\h+0.28) {\h};
    \node at (\x+0.325,-0.38) {\lab};
  }
\end{scope}
\begin{scope}[shift={(6.2,0)}]
  \node[anchor=west,font=\bfseries\small] at (0,8.8) {(b) Publication maturity by component};
  \draw[->,black!55] (0,0) -- (0,8.3);
  \draw[->,black!55] (0,0) -- (5.2,0);
  \fill[gray!25] (0.7,0) rectangle +(0.65,6); \fill[orange!35] (0.7,6) rectangle +(0.65,2);
  \fill[gray!25] (1.8,0) rectangle +(0.65,5); \fill[orange!35] (1.8,5) rectangle +(0.65,2);
  \fill[gray!25] (2.9,0) rectangle +(0.65,4);
  \fill[orange!35] (4.0,0) rectangle +(0.65,5); \fill[red!25] (4.0,5) rectangle +(0.65,1);
  \foreach \x/\lab in {0.7/C1,1.8/C2,2.9/C3,4.0/C4}{\node at (\x+0.325,-0.38) {\lab};}
  \fill[gray!25] (0.3,-1.0) rectangle +(0.35,0.25); \node[anchor=west] at (0.75,-0.88) {Archival};
  \fill[orange!35] (2.0,-1.0) rectangle +(0.35,0.25); \node[anchor=west] at (2.45,-0.88) {Preprint};
  \fill[red!25] (3.8,-1.0) rectangle +(0.35,0.25); \node[anchor=west] at (4.25,-0.88) {Non-arch.};
\end{scope}
\begin{scope}[shift={(12.4,0)}]
  \node[anchor=west,font=\bfseries\small] at (0,6.9) {(c) Focal-set composition};
  \draw[->,black!55] (0,0) -- (0,6.3);
  \draw[->,black!55] (0,0) -- (5.2,0);
  \fill[violet!28] (0.6,0) rectangle +(0.85,6.4); \draw[violet!65!black] (0.6,0) rectangle +(0.85,6.4);
  \fill[gray!30] (2.15,0) rectangle +(0.85,2.4); \draw[gray!65] (2.15,0) rectangle +(0.85,2.4);
  \fill[red!25] (3.7,0) rectangle +(0.85,1.2); \draw[red!60!black] (3.7,0) rectangle +(0.85,1.2);
  \node at (1.025,6.68) {16}; \node at (2.575,2.68) {6}; \node at (4.125,1.48) {3};
  \node[align=center] at (1.025,-0.55) {Core\\LLM/VLM};
  \node[align=center] at (2.575,-0.55) {Non-LLM\\diagnostic};
  \node[align=center] at (4.125,-0.55) {Boundary};
\end{scope}
\begin{scope}[shift={(18.6,0)}]
  \node[anchor=west,font=\bfseries\small] at (0,6.9) {(d) Component isolation};
  \draw[->,black!55] (0,0) -- (0,6.3);
  \draw[->,black!55] (0,0) -- (5.2,0);
  \fill[teal!35] (0.6,0) rectangle +(0.85,6.0); \draw[teal!70!black] (0.6,0) rectangle +(0.85,6.0);
  \fill[orange!35] (2.15,0) rectangle +(0.85,0.8); \draw[orange!70!black] (2.15,0) rectangle +(0.85,0.8);
  \fill[gray!30] (3.7,0) rectangle +(0.85,3.2); \draw[gray!65] (3.7,0) rectangle +(0.85,3.2);
  \node at (1.025,6.28) {15}; \node at (2.575,1.08) {2}; \node at (4.125,3.48) {8};
  \node[align=center] at (1.025,-0.55) {Dedicated};
  \node[align=center] at (2.575,-0.55) {Partial};
  \node[align=center] at (4.125,-0.55) {None};
\end{scope}
\end{tikzpicture}%
}
\caption{Evidence landscape of the 25 focal methods. Panel~(c) shows
the focal-set composition explicitly: 16 core LLM/VLM-augmented methods,
six non-LLM diagnostic/exemplar methods, and three planning/robotics
boundary cases. Counts use the retained primary assignment. Publication maturity and component isolation are coded as separate
dimensions and are not collapsed into a single quality score. Panel~(b) makes the lower maturity of C4 visible: its
six focal entries comprise five preprints and one non-archival report,
whereas C3 contains four archival studies. Machine-readable counts and
row-level coding are supplied with the supplementary data.}
\label{fig:evidence-landscape}
\end{figure*}
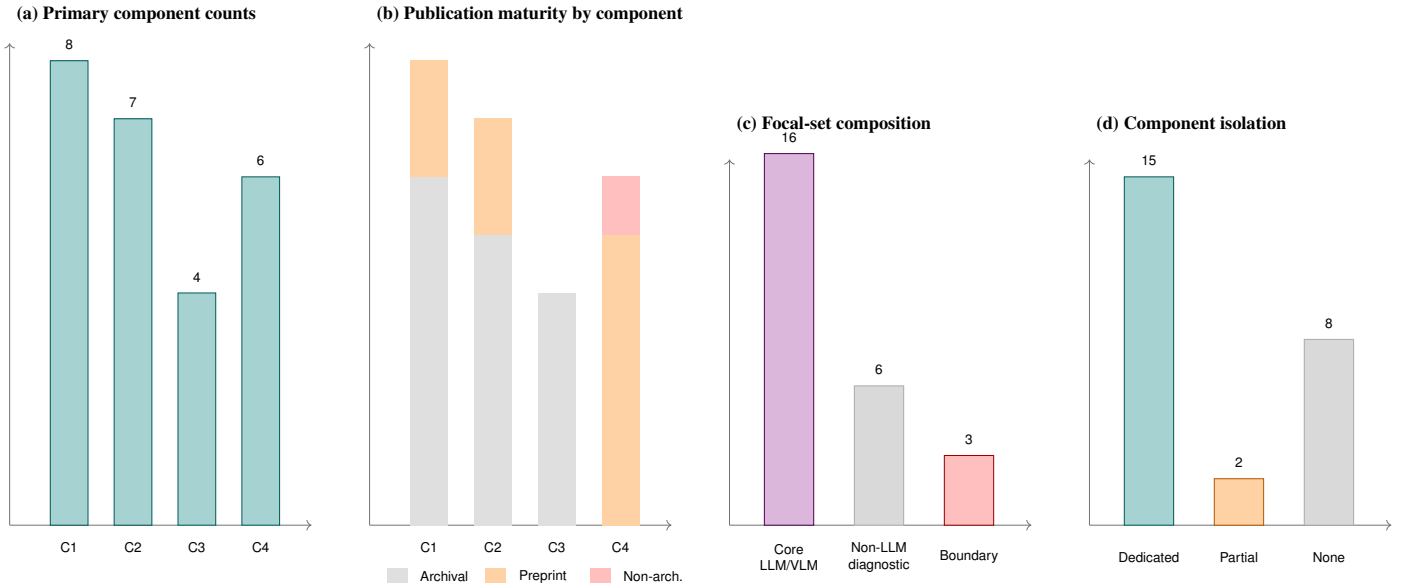

\noindent Figure~\ref{fig:evidence-landscape} shows a corpus-level
summary that complements the method tables. The numerical proximity of the component counts should not be read as
equal evidential maturity:
publication status and component isolation differ sharply across
components, especially for C4.

\section{C1: Context Construction}
\label{sec:c1}

\textbf{What C1 does.}
Before any anticipation step, the observed video prefix
$X_{1:t}$ is transformed into an enriched representation
$\tilde{X}$ that captures the past in a form suitable for
downstream goal reasoning (C2) and future decoding (C3).
Because all later components depend on the fidelity, scope, and
semantic form of this representation, C1 strongly conditions the
quality of downstream anticipation. In practice, missing,
compressed, or noisy context cannot usually be recovered by a
more sophisticated decoder alone.

\subsection{Design Factors}

\begin{table}[H]
\scriptsize\centering
\caption{Design factors in \textbf{C1: Context Construction}.
Trade-offs are grounded in cited ablation evidence.
HOI\,=\,hand-object interaction; MMR\,=\,maximal marginal relevance.}
\label{tab:c1-factors}
\setlength{\tabcolsep}{2pt}
\renewcommand{\arraystretch}{1.1}
\begin{tabularx}{\columnwidth}{p{1.1cm}Yp{1.25cm}p{1.85cm}}
\toprule
\textbf{Factor} & \textbf{Choices} & \textbf{Trade-off} & \textbf{Example} \\
\midrule
Modality & RGB vs.\ +flow/audio/HOI & Richer vs.\ heavier & M-CAT~\citep{beedu2024mcat}, PALM~\citep{kim2024palm} \\
Format & Labels / narrations / captions & Semantic vs.\ fast & TransFusion~\citep{pasca2024transfusion} \\
Scope & Window $N$ vs.\ full history & Rich vs.\ length limit & PALM ($N$ ablation) \\
Retrieval & None / MMR / similarity & Diverse vs.\ costly & PALM (MMR ICL) \\
Spatial & Global / HOI hotspots & Precise vs.\ context loss & AFF-ttention!~\citep{yang2024affttention} \\
Text role & Past only / future target / both & Strong vs.\ $2\times$ cost & M-CAT~\citep{beedu2024mcat} \\
\bottomrule
\end{tabularx}
\end{table}

\subsection{Comparative Evidence by Design Pattern}

\paragraph{Semantic compression and retrieval}
Jing et al.'s article ``Multimodal adaptive fusion for enhanced long-term
action anticipation'' introduces the Semantic-Guided Adaptive Fusion
Transformer (SAFT). Together with TransFusion and PALM, SAFT converts
observed activity into a more compact semantic history, but the three methods
make different trade-offs
\citep{pasca2024transfusion,kim2024palm,saft2026}. TransFusion separates
perceptual encoding from natural-language abstraction; PALM combines
recognised verb--noun histories, VLM captions, and diversity-aware
exemplar retrieval; SAFT replaces retrieval with an incremental textual
correction stream. These comparisons suggest that context format and
history selection can be as consequential as the downstream decoder,
while also showing that semantic compression inherits recogniser and
captioning errors.

\paragraph{Complementary modalities and adaptation}
AAG/AAG+ test whether depth and prior-action semantics can compensate
for sparse temporal observation, whereas DCPGN adapts prototype memory
and textual/visual clues at test time across ego--exo views
\citep{benavent2026aag,benavent2026aagplus,shi2026dcpg}. These studies
extend C1 beyond ``more frames versus fewer frames'': the relevant
comparison is between dense video and a controlled bundle of
complementary context, and between static representations and context
that adapts under domain shift.

\paragraph{Offline language supervision without a runtime LLM}
M-CAT uses LLM-derived text for observed-context conditioning and future
contrastive targets, but avoids a test-time language-model call
\citep{beedu2024mcat}. It therefore isolates a distinct design option:
language can shape representation learning without becoming part of the
runtime decoder. Reported long-tail gains are informative, but direct
comparison with visual systems still requires matched inference and
pretraining budgets.

\paragraph{Strong non-LLM representation controls}
The V-JEPA family and EgoAnticipator show why C1 must be controlled
before attributing a gain to language
\citep{assran2025vjepa2,murLabadia2026vjepa21,chen2025egoanticipator}.
Their dense predictive pretraining, probe design, loss, memory, and clip
sampling differ substantially from language-augmented systems. The
published EK-100 values therefore diagnose a configuration/evaluator
mismatch, not a clean ordering. EgoVideo and subsequent challenge
reports further reinforce the same qualitative point: encoder quality
strongly conditions the action history supplied to any later language
decoder~\citep{pei2024egovideo,chu2025ego4dlta,chu2026jfaa,wang2026tapjepa,chu2026vista,froststa2026}.

\noindent Across C1, the comparative finding is not that textual context
or visual pretraining is universally preferable. It is that semantic
compression, multimodal context, adaptation, and predictive pretraining
change the information available to every later component; claims about
C2 or C3 therefore require a controlled C1 baseline.

\subsection{Failure Modes}

\emph{ASR and detection noise:}
text-mediated pipelines are sensitive to transcription and
detection errors, which can degrade alignment between narration,
objects, and action chains. PALM explicitly reports this
sensitivity on EPIC-KITCHENS under varying ASR
quality~\citep{kim2024palm}. To make the mechanism concrete:
when background noise corrupts a narration transcript from
``take bowl'' to ``take ball'', the textual context becomes
misaligned with the actual action chain, and the predicted future
can drift toward a sequence appropriate for a sporting activity
rather than the omelette preparation actually under way.

\emph{Domain shift:}
vocabularies and object priors tuned to kitchen environments often
transfer poorly to other settings. Open-vocabulary detectors, such
as those used in ObjectPrompt~\citep{zhang2024objectprompt},
partially mitigate this problem but do not eliminate it; object
priors learned on EPIC-KITCHENS, for example, transfer poorly to
assembly settings such as Assembly101~\citep{sener2022assembly101}, where the interaction
vocabulary is dominated by tools and parts rather than utensils
and ingredients.

\emph{Context-length limits:}
long activity histories can exceed practical LLM context windows,
making compression and summarization mechanisms essential;
deployment-oriented compression strategies are discussed in E2
(Section~\ref{sec:deploy}). As an illustration of scale, a
twenty-minute cooking session mapped to per-action textual
descriptions can quickly saturate a 4096-token context window,
silently dropping earlier evidence (for instance, that the user
already made coffee and is unlikely to repeat a beverage step)
that would otherwise disambiguate the future.

\subsection{C1 in Third-Person and Procedural Settings}
\label{sec:c1-third}

The C1 design factors above are characterised primarily through
egocentric benchmarks such as Ego4D and EPIC-KITCHENS, but related
issues also arise in third-person and procedural settings with
different emphases. In egocentric video, manipulated objects are
typically prominent, hand--object interactions provide natural
spatial anchors, and narration-aligned textual cues may be
available. In third-person settings such as
50Salads~\citep{stein2013fiftysalads} and
Breakfast~\citep{kuehne2014breakfast}, objects are smaller, action
boundaries are often less sharply localised, and narration is
usually absent. Procedural benchmarks such as
COIN~\citep{tang2019coin}, CrossTask~\citep{zhukov2019crosstask}, and
Assembly101~\citep{sener2022assembly101} introduce an additional challenge:
errors in context construction can propagate across long
multi-step procedures with stronger ordering constraints.

A practical consequence is that \emph{context format}
(Table~\ref{tab:c1-factors}) becomes even more important outside
the egocentric setting. Raw label sequences obtained from a
third-person recogniser often carry weaker semantic fidelity than
egocentric narrations or interaction-focused summaries, which
increases the need for explicit captioning, step detection, or
other procedure-aware context builders. ActionLLM reports results
on 50Salads and Breakfast~\citep{wu2025actionllm},
VideoPlan on COIN and CrossTask~\citep{videoplan2026}, and
GP-AMS on Assembly101~\citep{gpams2026}. These results suggest that
parts of the C1 design logic transfer beyond
egocentric benchmarks, although the C1 bottleneck tends to become
more pronounced where upstream recognition is weaker. Because the
present survey is centred on Ego4D and EPIC-KITCHENS, this
cross-setting coverage remains secondary and is acknowledged as a
scope limitation in Section~\ref{sec:limitations}.

\paragraph{Bridge to C2}
The enriched representation $\tilde{X}$ produced by C1 feeds
directly into goal and intention modelling. An important interaction is
that the \emph{format} of the context can matter as much as its
raw informational content: a sequence of action labels alone is
often less goal-discriminative than a representation enriched
with object interactions, textual summaries, or procedure-aware
captions, because the same observed prefix may remain consistent
with several distinct high-level goals. The prefix
``take pan, heat oil'', for instance, is equally consistent with
omelette preparation, a pasta sauce, or a stir-fry. When the C1
output is semantically impoverished in this sense, C2 must rely
more heavily on language-model priors and less on video-grounded
evidence, increasing the risk of goal hallucination.
\section{C2: Goal and Intention Modelling}
\label{sec:c2}

\textbf{What C2 does.}
Given the enriched context $\tilde{X}$ produced by C1, C2 infers
the actor's latent goal $G$: the high-level behavioural objective
that makes some futures plausible and others unlikely. This
reduces ambiguity, improves long-horizon coherence, and provides a
more meaningful basis for downstream decoding in C3.

Goal and intention modelling is the component that most clearly
distinguishes recent LLM-augmented anticipation systems from
earlier pattern-matching approaches. A decoder that operates only
on observed action regularities can often predict the immediate
next step, but it is much weaker at modelling why the current
action chain is unfolding and which longer-range futures are
consistent with that underlying purpose. C2 therefore addresses a
distinct design problem: not merely predicting the next action,
but inferring the objective that organises the future sequence.

Because the terms goal, intention, subgoal, objective, and task
are used variably across the surveyed literature, and because the
distinctions among them matter for classifying methods correctly,
we fix the following operational vocabulary before analysing the
design space.

\begin{conceptbox}
These terms are used variably across the literature. We adopt the
following operational distinctions throughout this survey.

\textbf{Intention:} the inferred latent \emph{behavioural
objective} that motivates the observed action chain, that is, the
``why'' of the activity. Example: ``the actor intends to prepare a
meal.'' Intention is typically implicit, must be inferred from
context, and is the broadest and most abstract of the three
levels.

\textbf{Goal} ($G$): the \emph{operational target} that directly
conditions future action prediction, that is, the ``what'' the
actor is working toward in the near term. Example: ``the goal is
to make an omelette.'' A goal is more specific than an intention
and functions as the conditioning variable in
Eq.~\eqref{eq:goal}.

\textbf{Subgoal:} an \emph{intermediate procedural state} on the
path to the goal, that is, the ``next milestone.'' Example:
``crack three eggs into the bowl before whisking.'' Subgoals are
operationalised differently across methods: as step-level targets
in AntGPT~\citep{zhao2024antgpt}, as PDDL sub-conditions in
Anticipate \& Act~\citep{arora2025anticipate}, and as latent MPC
waypoints in HWM~\citep{zhang2026hwm}.

These distinctions matter for taxonomy: INSIGHT operates primarily
at the intention level, AntGPT at the goal level, and HWM at the
subgoal level. Blurring the three concepts produces taxonomies in
which methods with qualitatively different architectures are
classified together.
\end{conceptbox}

The C2 design space varies primarily along six related questions:
(i)~how the goal is represented, whether as a latent variable,
textual intention, or executable task description;
(ii)~how it is inferred from the observed video context;
(iii)~how uncertainty in that inference is handled;
(iv)~how tightly the inferred goal is coupled to the downstream
decoder;
(v)~at what temporal granularity the goal is represented; and
(vi)~what supervision signal supports intention learning. In some
methods, the inferred goal is local and step-specific, whereas in
others it is represented at the level of subgoals or full
procedures. Table~\ref{tab:c2-factors} summarises the main design
factors.

\subsection{Design Factors}

\begin{table}[H]
\footnotesize\centering
\caption{Design factors in \textbf{C2: Goal/Intention Modelling}.
MAP\,=\,maximum a posteriori.}
\label{tab:c2-factors}
\setlength{\tabcolsep}{2pt}
\renewcommand{\arraystretch}{1.1}
\begin{tabularx}{\columnwidth}{p{1.4cm}Yp{1.55cm}p{1.35cm}}
\toprule
\textbf{Factor} & \textbf{Choices} & \textbf{Trade-off} & \textbf{Example} \\
\midrule
Goal representation & Latent state / textual intention / executable goal & Flexible vs.\ interpretable/executable & AntGPT, Ant.\&Act \\
Inference pathway & Prompted LLM / VLM$\rightarrow$text$\rightarrow$LLM / explicit reasoning policy & Simplicity vs.\ visual grounding & AntGPT, ICVL, INSIGHT \\
Uncertainty handling & MAP goal / top-$K$ goals / posterior-conditioned decoding & Efficient vs.\ robust & AntGPT, Mascar\'o et al. \\
Coupling to decoder & Prompt-level conditioning / feature-level fusion / constrained decoding & Modular vs.\ tightly aligned & ICVL, AntGPT \\
Temporal granularity & Immediate next goal / multi-step subgoal / procedure-level objective & Local precision vs.\ long-horizon coherence & GP-AMS \\
Supervision source & Implicit anticipation loss / auxiliary intention supervision / structured task signal & Broad applicability vs.\ stronger bias & Mascar\'o et al., GP-AMS \\
\bottomrule
\end{tabularx}
\end{table}

\subsection{Comparative Evidence by Design Pattern}

\paragraph{Explicit goal inference versus visually grounded intention}
AntGPT separates top-down goal inference from bottom-up sequence
completion, while ICVL first predicts and fuses a visually grounded
textual intention before language decoding
\citep{zhao2024antgpt,cao2025icvl}. The comparison clarifies the central
C2 trade-off: explicit procedural priors can constrain a long future,
but an intermediate goal estimate becomes an additional error channel.
Neither study isolates the language model from all representation and
decoder changes, so their ablations support the direction of a goal-conditioning effect rather than a universal LLM effect.

\paragraph{Semantic anchors and trainable reasoning}
PAR-VLA constructs verb and noun prototypes in an aligned
vision--language space and performs closed-set retrieval without
runtime text generation; INSIGHT gives intention reasoning its own
optimisation path; GP-AMS transfers goal-guided prompting and adaptive
modality selection to assembly activity
\citep{shao2026parvla,insight2026,gpams2026}. Together they show that C2
is broader than prompting: the common design object is a semantic or
intent state that changes future prediction, regardless of whether it
is produced by an autoregressive LLM.

\paragraph{Non-LLM intent controls}
Mascar\'o et al. model latent intention before the recent LLM wave, and
TrajPilot uses predicted camera trajectories as a geometric intent
proxy in an action-aligned space
\citep{mascaro2023intention,jun2026trajpilot}. These controls prevent
``goal reasoning'' from being equated with language-model use. In the
reported TrajPilot comparisons, geometric conditioning can outperform
textual conditioning in the studied regimes, motivating direct tests of
which intent representation is informative rather than assuming that
language is the privileged interface.

\noindent The cross-method C2 finding is therefore conditional: explicit
or semantic goal information can help when it is reliable and when
multiple futures remain plausible, but current systems generally commit
to one goal estimate and rarely propagate goal uncertainty into C3.

\subsection{Failure Modes}

\emph{Goal ambiguity propagation:}
when the observed context is semantically weak or incomplete,
multiple high-level goals remain plausible. In such cases,
goal-conditioned prediction can amplify ambiguity rather than
resolve it, because the decoder becomes sensitive to whichever
goal hypothesis is selected upstream. Consider the observed
prefix ``take bowl, add flour, stir'': this is equally consistent
with pancakes, bread dough, and b\'echamel sauce. A MAP goal
inference commits to one; if that commitment is wrong, the decoder
may generate a coherent twenty-step pancake sequence when the
ground truth is bread, producing high edit distance at long
horizons even when the first few predicted steps are correct.

\emph{Goal hallucination under weak evidence:}
LLM-based goal inference may fall back on dataset priors,
commonsense continuations, or frequent procedural patterns that
are only weakly supported by the actual video evidence. This is
especially problematic when the context window is short or the C1
representation has been aggressively compressed, a pattern already
hinted at by context-reduction ablations in
AntGPT~\citep{zhao2024antgpt}: reducing observed context from
$N{=}8$ to $N{=}1$ segments degrades performance substantially,
suggesting that goals inferred from short contexts reflect
language-model priors more than video evidence.

\emph{Coupling mismatch:}
if the inferred goal is injected only weakly into the decoder, its
effect may be overridden by the decoder's own sequence prior; if
it is imposed too strongly, errors in goal inference can
overconstrain decoding and suppress valid alternatives. Existing
ablations in AntGPT and ICVL suggest that this trade-off is not
merely theoretical, but already visible in current
designs~\citep{zhao2024antgpt,cao2025icvl}: the top-down goal
path helps when context is rich but can degrade performance when
context is short, precisely because an unreliable goal imposes
stronger constraints than no goal at all.

\emph{Evaluation mismatch:}
current anticipation benchmarks primarily evaluate the predicted
action sequence rather than the correctness of the inferred goal
itself. As a result, a method may achieve competitive next-action
or sequence metrics even when its intention reasoning is only
partially correct or only indirectly useful.

These failure modes suggest that C2 is most useful when goal inference
is grounded in the observed video and coupled strongly enough to
influence decoding, but not so strongly that early goal errors drive the
predicted sequence.

\paragraph{Bridge to C3}
The goal distribution $P(G \mid \tilde{X})$ inferred in C2 helps
determine which futures C3 should generate. An important
cross-component interaction is the strength of this conditioning.
When goal information is injected only weakly, its effect may be
limited by the decoder's own distributional prior; when imposed
too strongly, errors in goal inference can overconstrain decoding
and reduce recall. Existing ablations in methods such as AntGPT
and ICVL suggest that this coupling--accuracy trade-off is real,
but a systematic characterisation across datasets and prediction
horizons remains absent. The C2--C3 interface therefore embodies
an implicit bias--variance trade-off that neither component
captures alone: a MAP goal reduces decoding ambiguity but
introduces a single point of failure, while fuller marginalisation
over goals (Eq.~\eqref{eq:goal}) may be more robust but can be
computationally costly for autoregressive LLM decoders.
\section{C3: Future Prediction and Decoding}
\label{sec:c3}

\textbf{What C3 does.}
Given the enriched context $\tilde{X}$ and, where available, the
goal representation $G$, C3 generates the predicted future action
sequence $\hat{Y}_{t+1:T}$. This is the component in which the
model must convert context and intention into an explicit future:
a plausible next action and a temporally coherent sequence over the
anticipation horizon. C3 must determine both \emph{what} to predict and
\emph{how} to decode futures without collapsing into repetition,
drifting off-manifold, or violating simple procedural structure.

C3 inherits several well-studied problems from the language
modelling literature, including hallucinated
generation~\citep{ji2023hallucination} and exposure bias in
sequence-to-sequence training~\citep{ranzato2015sequence}, for
which remedies such as scheduled
sampling~\citep{bengio2015scheduled} were originally developed.
Three aspects, however, become especially consequential in video
anticipation. First, \emph{temporal causality}: unlike free text,
predicted actions must respect a physical timeline; an actor
cannot ``add egg'' before ``crack egg'', yet autoregressive
decoders carry no explicit mechanism to enforce such ordering.
Second, \emph{physical consistency}: whereas a language model's
hallucination produces merely implausible text, an anticipation
model's hallucination produces physically impossible action
sequences, such as stirring an ingredient that is not present in
the scene, and such errors are harder to detect automatically than
textual ones. Third, \emph{multimodal grounding}: exposure bias is
compounded in the video setting because early prediction errors
sever the decoder's connection to visual evidence; once decoding
drifts, there is no mechanism to re-anchor the generated future to
the observed scene. These three difficulties explain why C3
designs in the surveyed literature invest so heavily in
plausibility control, multi-token prediction, and cross-modal
anchoring, respectively.

The C3 design space varies primarily along five related questions:
(i)~what output space is predicted, whether closed-set action
labels or more open textual forms;
(ii)~how future tokens are decoded, for example greedily, through
beam-style search, or under stochastic sampling;
(iii)~whether futures are generated one step at a time or through
multi-token prediction;
(iv)~how plausibility and repetition are controlled during
training or decoding; and
(v)~what auxiliary objectives, if any, are used to stabilise
long-horizon sequence prediction. Table~\ref{tab:c3-factors}
summarises the main design factors.

\subsection{Design Factors}

\begin{table}[H]
\footnotesize\centering
\caption{Design factors in \textbf{C3: Future Prediction \& Decoding}.
MTP\,=\,multi-token prediction; ED\,=\,edit distance.}
\label{tab:c3-factors}
\setlength{\tabcolsep}{2pt}
\renewcommand{\arraystretch}{1.1}
\begin{tabular}{p{1.5cm}p{2.1cm}p{1.7cm}p{1.45cm}}
\toprule
\textbf{Factor} & \textbf{Choices} & \textbf{Trade-off} & \textbf{Example} \\
\midrule
Output space & Closed labels / open text / hybrid verbalisation & Unambiguous vs.\ expressive & AntGPT, PALM \\
Decoding policy & Greedy / beam / temperature-based sampling & Stable vs.\ diverse & AntGPT, PlausiVL \\
Sequence granularity & Stepwise autoregressive / multi-token prediction & Flexible vs.\ globally coherent & VideoPlan \\
Plausibility control & None / penalty / compatibility or counterfactual constraints & Simple vs.\ constrained & PlausiVL \\
Auxiliary training signal & Anticipation only / +auxiliary tasks / +consistency losses & Cleaner objective vs.\ richer supervision & VideoPlan, PlausiVL \\
Label normalisation & Native closed-set output / post-hoc text-to-label mapping & Direct evaluation vs.\ mapping noise & ActionLLM, PALM \\
\bottomrule
\end{tabular}
\end{table}

\subsection{Comparative Evidence by Design Pattern}

\paragraph{Sequence stability and planning-oriented objectives}
PlausiVL targets repetition and verb--noun plausibility directly,
whereas VideoPlan supervises structured chunks of the future through
auxiliary tasks and multi-token prediction
\citep{mittal2024plausivl,videoplan2026}. Their common lesson is that
long-horizon error depends on model capacity, objective design, and
decoding constraints, which determine whether local predictions compose
into a coherent sequence. Multi-token prediction and scheduled sampling
remain distinct mechanisms, addressing output granularity and exposure
distribution respectively~\citep{bengio2015scheduled}.

\paragraph{Visual anchoring and explicit action dependencies}
ActionLLM maintains visual--text interaction inside a largely frozen
language decoder, while the non-LLM AGA attends over predicted action
distributions rather than raw appearance
\citep{wu2025actionllm,tai2026aga}. AGA is an especially important
control: semantic dependency reasoning can improve future decoding
without a language model, so gains from C3 should be attributed to the
specific dependency or stability mechanism rather than to model family
alone.

\paragraph{Broader decoder evidence}
BiAnt, ALM, and LLMAction extend language-based sequence modelling;
Action Sequence Augmentation and VEDIT provide non-LLM controls based on
grammar-aware augmentation and latent diffusion; VideoLLM is retained as a
historical precursor to prediction-as-language
\citep{sato2025biant,li2024alm,wang2025llmaction,qiu2025asa,lin2025vedit,chen2023videollm}.
These records broaden the set of decoder designs without changing the focal
C3 factors.

\noindent Across C3, the recurring comparative finding is that sequence
stability, uncertainty, and visual anchoring must be evaluated
separately from the language prior. A stronger decoder can compensate
for some C1/C2 errors, but it can also amplify an incorrect context or
goal over the full horizon.

\subsection{Failure Modes}

\emph{Hallucinated futures:}
decoders may generate action sequences that are linguistically or
procedurally plausible yet insufficiently supported by the visual
context, especially under ambiguity or domain
shift~\citep{mittal2024plausivl}. A concrete instance: a system
observing ``take bowl, crack egg, whisk'' may confidently predict
``add milk, add sugar, pour batter, cook pancake'' even when the
visual scene contains no pancake ingredients at all; the sequence
is plausible under the language model's training distribution but
unsupported by the video.

\emph{Long-horizon repetition and collapse:}
autoregressive decoding can drift toward a small set of frequent
actions, producing repetitive or low-information futures. On
twenty-step Ego4D-LTA sequences ($Z{=}20$), unconstrained decoders
often converge to alternating ``take object, put object'' pairs.
PlausiVL shows that explicit anti-repetition mechanisms can partly
mitigate this effect~\citep{mittal2024plausivl}.

\emph{Exposure bias across the horizon:}
when each decoded step conditions on previous predictions, early
errors can propagate and distort later parts of the sequence.
This problem becomes more severe as the prediction horizon grows
and motivates multi-token or planning-oriented alternatives such
as VideoPlan~\citep{videoplan2026}.

\emph{Normalisation artefacts:}
when open-vocabulary or free-text outputs must be mapped back to a
closed benchmark label space, post-hoc canonicalisation errors can
distort evaluation, sometimes inflating edit distance even when
the generated future is semantically close to the ground truth.
When a decoder generates ``pick up the mixing bowl'' as free text,
for example, mapping this output to the EPIC-KITCHENS label
\texttt{take:bowl} requires a normalisation step whose errors are
charged against the anticipation model.

Strong C3 design requires a capable decoder and explicit mechanisms for
keeping generated futures stable, non-repetitive, and evaluable under
benchmark constraints.

\paragraph{Bridge to C4}
C3 produces one or more candidate future sequences $Y$, but even
goal-conditioned and fluent outputs may still violate logical,
physical, or task-specific constraints that token-level decoding
does not reliably enforce. This motivates C4, which introduces
external grounding or executability checks over the generated
future. In some settings, lightweight symbolic constraints can
prune invalid candidates early and thereby improve both quality
and efficiency; in others, especially when validation requires
simulator access or costly world models, the additional grounding
stage introduces its own inference burden. The practical
trade-off therefore depends on how expensive constraint checking
is relative to candidate generation.
\section{C4: Emerging Grounding and Executability Extensions}
\label{sec:c4}

\textbf{Evidence status and function.}
C4 is retained as an analytically important but less mature extension
of the core anticipation pipeline. Its evidence base mixes standard
anticipation benchmarks with recent preprints, a non-archival thesis,
and planning or robotics boundary cases. Accordingly, C4 supports
mechanism-level hypotheses and reporting recommendations more strongly
than general performance claims.

Even strong decoders can produce fluent but world-inconsistent
predictions, such as action sequences that violate object
affordances, temporal preconditions, or basic physical
executability. C4 therefore constrains decoding to a feasible set
$\mathcal{F}(X_{1:t}, S)$ induced by an external grounding state
$S$, for example a symbolic program state, a geometric scene
representation, or an embodied dynamics model. The role of C4 is
not to replace C3, but to rule out futures that remain
linguistically plausible yet operationally invalid.

Before analysing methods, it is worth separating three notions
that frequently blur together in the surveyed literature. A
predicted future is \emph{plausible} when it is statistically
likely given the training distribution; this is a C3 property. It
is \emph{consistent} when it does not violate the observed scene
state, for example by predicting ``use knife'' when no knife is
present; this is a C4 property requiring scene grounding. It is
\emph{executable} when it is physically and procedurally feasible
for an agent to carry out in sequence; this is a stronger C4
property that typically requires world models or planners. A
prediction can be plausible without being consistent, and
consistent without being executable, and C4 methods differ
primarily in which of these levels of checkability they target.

The central design question in C4 is how much structure should be
imposed beyond token-level decoding. Some methods use explicit
symbolic constraints and planners, others rely on scene or
trajectory grounding, and still others push toward embodied
world-model planning. These approaches differ in how they
represent feasibility, how costly that feasibility check is, and
how tightly it is coupled to the anticipation model. In this
sense, C4 is best understood as the component that turns
``plausible'' futures into futures that are at least partially
\emph{checkable}. Table~\ref{tab:c4-factors} summarises the main
design factors.

\subsection{Design Factors}

Table~\ref{tab:c4-factors} sets out the axes along which C4 designs
differ. Two of them deserve emphasis before the methods are
discussed. \emph{Constraint timing} determines whether feasibility
is checked after decoding, imposed during it, or folded into joint
planning, and this largely fixes how expensive the check is.
\emph{Transfer burden} captures what must be rebuilt when the domain
changes, which is the practical limit on every symbolic approach
reviewed below.

\begin{table*}[t]
\footnotesize\centering
\caption{Design factors in \textbf{C4: Grounding \& Executability}.
KG\,=\,knowledge graph; MPC\,=\,model predictive control.}
\label{tab:c4-factors}
\setlength{\tabcolsep}{5pt}
\renewcommand{\arraystretch}{1.15}
\begin{tabularx}{\textwidth}{
>{\raggedright\arraybackslash}p{2.2cm}
>{\raggedright\arraybackslash}X
>{\raggedright\arraybackslash}p{3.0cm}
>{\raggedright\arraybackslash}p{2.8cm}}
\toprule
\textbf{Factor} & \textbf{Choices} & \textbf{Trade-off} & \textbf{Example} \\
\midrule
Grounding state & Symbolic program / scene graph / trajectory or latent dynamics
& Interpretable vs.\ expressive
& LEAP, SymAnt, HWM \\

Constraint source & Hand-crafted / learned / hybrid
& Reliable vs.\ scalable
& SymAnt, TR-LLM \\

Constraint timing & Post-hoc filtering / decoder-time conditioning / joint planning
& Modular vs.\ tightly coupled
& PlausiVL, LEAP, Ant.\&Act \\

Planner type & None / symbolic planner / hierarchical MPC
& Simplicity vs.\ stronger executability
& Ant.\&Act, HWM \\

Evaluation target & Standard anticipation metrics / logic or solvability checks / task success
& Comparable vs.\ realistic
& LEAP, Ant.\&Act, HWM \\

Transfer burden & Domain-specific schemas / reusable priors / embodied adaptation
& Strong control vs.\ portability
& SymAnt, TR-LLM \\
\bottomrule
\end{tabularx}
\end{table*}

\subsection{Comparative Evidence by Design Pattern}

\paragraph{Symbolic programmes and verification}
LEAP encodes futures as action programmes; FactCheck adds an
Observer--Planner--Verifier loop over object state, affordance, and
preconditions; and the non-archival SymAnt combines scene and knowledge
graphs
\citep{dessalene2023leap,cao2026factcheck,aryan2025symant}. These methods
make futures more checkable, but also expose a controllability--coverage
trade-off. FactCheck's verifier can suppress unusual yet correct
futures, while graph and programme approaches inherit ontology and
schema maintenance costs. SymAnt is therefore retained as a design
exemplar, not as mature evidence of C4 effectiveness.

\paragraph{Geometric grounding}
TR-LLM constrains semantic prediction with human-trajectory information
\citep{trllm2025}. It shows that C4 need not be symbolic: geometric
priors can narrow feasible futures without a hand-built action schema,
although they transfer the dependency to trajectory estimation and
scene coverage.

\paragraph{The planning boundary}
Anticipate \& Act converts predicted tasks into PDDL goals, and HWM uses
hierarchical latent-world-model planning
\citep{arora2025anticipate,zhang2026hwm}. These boundary cases clarify
what changes when an anticipation output must be executable: label
agreement is no longer sufficient, and solvability, task success, or
control cost become part of the evaluation target.

\noindent The comparative C4 hypothesis is that additional structure can
exclude invalid futures, but its value depends on transfer burden,
coverage loss, and the cost of search, simulation, or world-model
rollouts. Because task definitions and metrics differ sharply, C4 is
presented as an adjacent emerging extension and is not treated as
evidence-equivalent to C1--C3.

\subsection{Failure Modes}

\emph{Schema brittleness:}
symbolic and graph-based methods depend on curated ontologies,
preconditions, or affordance structures that may not transfer
cleanly across domains, object vocabularies, or task regimes. A
knowledge graph built for a kitchen, in which ``take mug'' is
typically followed by ``add milk'', returns misleading
preconditions in an office setting where the natural continuation
is ``use coffee machine'', causing the grounding stage to reject
valid futures.

\emph{Grounding incompleteness:}
the external state used for feasibility checking may itself be
partial or noisy. If the scene graph is wrong, the object state is
missing, or the trajectory prior is poorly estimated, the
grounding stage can reject valid futures or preserve invalid ones;
a detector that misclassifies a strainer as a bowl, for example,
propagates its error into every downstream trajectory constraint.

\emph{Subgoal reachability:}
in hierarchical or embodied settings, a high-level planner may
generate subgoals that are semantically sensible but not reachable
by the lower-level dynamics or controller. This is one of the main
failure patterns in world-model-based planning.

\emph{Metric mismatch:}
standard anticipation metrics such as ED or sequence-level scores
do not fully capture executability, logical consistency, or plan
solvability. A system could, in principle, achieve strong
Action@1s by predicting frequent actions while generating
logically invalid programs. As a result, C4 improvements may be
undervalued when evaluation remains restricted to label-sequence
agreement alone.
These failure modes motivate testing C4 when consistency or
executability is an explicit requirement. The current
evidence is not sufficient to prescribe C4 universally, and any gain
should be reported together with schema transfer, latency, and failure
coverage.

\paragraph{Bridge to E2/E3}
Because C4 often introduces external reasoning, search, or
validation, it interacts directly with deployment concerns.
Lightweight symbolic or graph-based checks can sometimes prune
invalid candidates early and reduce wasted decoding effort, while
heavier planners or world-model rollouts may instead add
substantial overhead. C4 is therefore both a correctness component and a
deployment-sensitive one: the practical
value of grounding depends on how its feasibility checks trade off
against latency, memory, and control requirements in the target
setting.
\section{Benchmarks and Protocol Comparability}
\label{sec:benchmarks}

Rigorous evaluation of action anticipation systems requires
careful attention to protocol details. Results reported under
incompatible settings are not directly comparable, yet such
comparisons are often made implicitly when methods differ in
backbone, modality, label space, context length, horizon, or
decoding setup.

To keep protocol awareness separate from raw leaderboard
aggregation, we adopt a two-level reporting scheme. The main paper
reports only a small set of clearly interpretable representative
comparisons (Tables~\ref{tab:ego4d} and~\ref{tab:epic}), while
Supplementary Table~S2 provides a broader evidence matrix
recording encoder, modality, label space, anticipation setting,
decoding setup, source provenance, and the achieved comparability
level. It distinguishes source-described alignment, code verification, and
independent reproduction. Comparability is not treated as a binary
property. Table~\ref{tab:compat-matrix}
additionally summarises, for each benchmark configuration, which
protocol conditions govern whether two reported results can be
compared at all.

\begin{metricbox}
\textbf{Ego4D LTA.}
Given $N$ observed segments, the task is to predict $Z$ future
verb--noun pairs and evaluate over $K$ candidates.
The \emph{official primary metric} is minimum edit distance at
$Z{=}20$ (ED~$\downarrow$), the minimum Levenshtein distance between
the ground-truth future sequence and the candidate predictions. Some
subsequent studies additionally report a sequence-level F1
(seq-F1~$\uparrow$); we treat this as a supplementary metric, and
where it is used its implementation and provenance should be stated
explicitly, since we could not identify an official challenge
implementation. Ego4D v1 and v2 use different label spaces, so
results are comparable only within the same version.

\textbf{EPIC-KITCHENS.}
Action anticipation is commonly evaluated at $\tau_a{=}1$s.
EK-55 typically reports recall-style mAP over
All/Frequent/Rare splits. The \emph{official} EK-100 anticipation
metric is \textbf{Mean Top-5 Recall}
(MT5R~$\uparrow$)~\citep{damen2022epic100,furnari2018leveraging}, also
written as class-mean Top-5 Recall. MT5R is computed by first
measuring, for each class $c$ appearing in the evaluation set, the
proportion of instances of $c$ whose ground-truth label falls within
the model's top-5 predictions, and then averaging these per-class
recalls over all such classes. It is reported separately for verb,
noun, and action, where an action prediction is correct only if the
verb and noun are jointly correct.

MT5R must not be confused with instance-level (micro-averaged) Top-5
recall, in which every test sample contributes equally. Because EK-100
has a pronounced long-tail class distribution, the two quantities can
differ substantially, and instance-level recall weights frequent classes more heavily and can
therefore yield more favourable scores for models whose predictions
concentrate on frequent verb--noun pairs. Throughout this survey we therefore write MT5R
rather than the ambiguous ``R@5'' whenever the official EK-100
protocol is meant, and we flag any cited number that we could not
confirm to be class-mean.

\textbf{Protocol caveat.}
Cross-paper comparisons are interpretable only when backbone,
pretraining data, modality, label space, evaluation subset (overall
validation, unseen participants, tail classes, or hidden test),
$N$, $Z$, $K$, and decoding policy are sufficiently aligned.
\end{metricbox}

\begin{table*}[t]
\footnotesize\centering
\caption{Source-verified subset of reported \textbf{Ego4D-LTA}
minimum edit-distance results. Values are transcribed from each
method's primary paper and are reported as separate verb, noun, and
action ED at $Z{=}20$; lower is better. All rows use $K{=}5$ and the
test split, but v1 and v2 label spaces must still be interpreted
separately. This table is intentionally selective rather than a
complete leaderboard: a row is included only when version, observed
context, horizon, candidate count, split, the full ED triple, and a
primary-source table could be confirmed. The values were not
independently reproduced.}
\label{tab:ego4d}
\setlength{\tabcolsep}{3pt}
\renewcommand{\arraystretch}{1.18}
\begin{tabular*}{\textwidth}{@{\extracolsep{\fill}}lcccccccl@{}}
\toprule
\textbf{Method} & \textbf{Ver.} & \textbf{$N$} & \textbf{$Z$}
  & \textbf{$K$} & \textbf{Verb ED} & \textbf{Noun ED}
  & \textbf{Action ED} & \textbf{Primary-source location} \\
\midrule
Mascar\'o et al.~\citep{mascaro2023intention}
  & v1 & 6 & 20 & 5 & 0.741 & 0.739 & 0.930 & Table~1; test set \\
AntGPT~\citep{zhao2024antgpt}
  & v1 & 8 & 20 & 5 & 0.6584 & 0.6546 & 0.8814 & Table~6; test set \\
PALM~\citep{kim2024palm}
  & v1 & 8 & 20 & 5 & 0.6559 & 0.6401 & 0.8613 & Table~1; test set \\
AntGPT~\citep{zhao2024antgpt}
  & v2 & 8 & 20 & 5 & 0.6503 & 0.6498 & 0.8770 & Table~6; test set \\
PALM~\citep{kim2024palm}
  & v2 & 8 & 20 & 5 & 0.6471 & 0.6117 & 0.8503 & Table~1; test set \\
\bottomrule
\end{tabular*}
\end{table*}

\begin{table*}[t]
\footnotesize\centering
\caption{Published EK-100 figures under nominally aligned
protocol descriptions. Values are reported as action Mean Top-5 Recall
(class-mean) on the overall validation split at $\tau_a{=}1$s and are
transcribed from the cited comparison sources. This is a
reporting-incompatibility audit, not a verified leaderboard: the rows attain
only Level~(i), prior baselines were not stated to have been rerun
under shared code, the survey did not re-execute them, and pretraining
is unmatched. Full provenance and exclusions are given in
Supplementary Table~S2.}
\label{tab:epic}
\setlength{\tabcolsep}{4pt}
\renewcommand{\arraystretch}{1.22}
\begin{tabularx}{\textwidth}{lccYccl}
\toprule
\textbf{Method} & \textbf{Year} & \textbf{Model / encoder config.}
  & \textbf{Family} & \textbf{Action MT5R} & \textbf{Publication status}
  & \textbf{Comparability} \\
\midrule
InAViT~\citep{roy2024inavit}
  & 2024 & 160M & Non-LLM, supervised & 25.8 & Archival & Level (i) \\
Video-LLaMA~\citep{zhang2023videollama}
  & 2023 & 7B & LLM-augmented & 26.0 & Archival & Level (i) \\
PlausiVL~\citep{mittal2024plausivl}
  & 2024 & 8B (Q-Former + LLM) & LLM-augmented & 27.6 & Archival & Level (i) \\
\midrule
V-JEPA~2~\citep{assran2025vjepa2}
  & 2025 & ViT-g384 (1B, frozen) & Non-LLM, dense SSL & 39.7 & Preprint$^\dagger$ & Level (i) \\
V-JEPA~2.1~\citep{murLabadia2026vjepa21}
  & 2026 & ViT-G (2B, frozen) & Non-LLM, dense SSL & 40.8 & Preprint$^\dagger$ & Level (i) \\
\bottomrule
\multicolumn{7}{l}{$^\dagger$ Preprint at the time of writing; interpret as emerging rather than settled evidence.} \\
\multicolumn{7}{l}{\parbox{0.95\textwidth}{\footnotesize\textit{Note:} Results on the
unseen-participants and tail-class subsets of EK-100, and results
reported as instance-level rather than class-mean recall, are
\emph{not} comparable to this table and are recorded separately in
Supplementary Table~S2.}} \\
\end{tabularx}
\end{table*}

\begin{table}[t]
\footnotesize\centering
\caption{Benchmark compatibility matrix. \checkmark~marks a protocol
condition that must be matched before two reported results can be
compared; ``fixed'' indicates the condition is fixed by the benchmark
definition. Subset refers to evaluation subsets such as overall
validation versus unseen participants. \textbf{Two uses must be
distinguished.} For \emph{leaderboard} comparison, methods may differ
in modality: using flow or object features is part of what is being
compared. For \emph{causal} attribution of a gain to one component,
modality must be held fixed or explicitly controlled, since otherwise
the modality budget confounds the component under test. The marks
below state the stricter, causal requirement; a leaderboard reader may
relax the modality column.}
\label{tab:compat-matrix}
\setlength{\tabcolsep}{2.5pt}
\renewcommand{\arraystretch}{1.2}
\begin{tabular}{lcccccc}
\toprule
\textbf{Benchmark} & \textbf{Metric} & \textbf{Horizon}
  & \textbf{Label space} & \textbf{Subset} & \textbf{Modality} \\
\midrule
Ego4D LTA v1  & ED / seq-F1 & \checkmark & v1 & --- & \checkmark \\
Ego4D LTA v2  & ED / seq-F1 & \checkmark & v2 & --- & \checkmark \\
EK-100 (official) & action MT5R & fixed & \checkmark & \checkmark & \checkmark \\
EK-100 (@1s)  & Action@1s & fixed & \checkmark & \checkmark & \checkmark \\
EK-55         & mAP & fixed & \checkmark & \checkmark & \checkmark \\
Assembly101   & task-specific & \checkmark & \checkmark & --- & \checkmark \\
\bottomrule
\end{tabular}
\end{table}

\subsection{Reconciling Reported EK-100 Numbers}
\label{sec:scale-discrepancy}

Assembling Table~\ref{tab:epic} surfaced a reporting pattern that we
record because it bears on how the table should be read.

Published results attributed to action Mean Top-5 Recall on the
EPIC-KITCHENS-100 validation set at $\tau_a{=}1$s span a wide range.
One line of work reports RULSTM at 14.0, AVT+ at 15.9, RAFTformer-2B
at 19.1, S-GEAR-2B at 19.5, and PAR-VLA~\citep{shao2026parvla} at
22.5 with RGB and 24.1 with RGB and flow, the last presented as state
of the art. The comparison table of
Assran et al.~\citep{assran2025vjepa2} reports InAViT at 25.8,
Video-LLaMA at 26.0, PlausiVL at 27.6, and V-JEPA~2 at 39.7, and the
JFAA challenge report~\citep{chu2026jfaa} independently reports a
figure close to 39.6 on validation.

We are explicit about what this does and does not show. It does
\emph{not} show that the two groups use different metrics: the upper
range is corroborated by an independent challenge report, PlausiVL's
own paper states its figure as class-mean Top-5 Recall, and large
gains from heavily pretrained encoders are not implausible on their
face. What it does show is that a reader cannot safely place a number
from one group beside a number from the other, since a 2026 paper
presents 24.1 as best reported under the cited split and metric while figures near 40 are reported
elsewhere for the same nominal protocol. The magnitude of this discrepancy
is plausibly explained by some combination of pretraining exposure, probe
design, annotation or subset selection, and evaluator implementation.
However, the available source descriptions do not establish which factor is
dominant; we therefore treat the record as an unresolved reporting
incompatibility rather than as a performance contradiction.

The practical consequence is a distinction we adopt throughout
between three levels of comparability: \textbf{(i)}~matched according
to the descriptions given in the source papers; \textbf{(ii)}~matched
using the same released evaluation code; and
\textbf{(iii)}~independently reproduced under identical data and
code. Table~\ref{tab:epic} attains level~(i) only. Its rows are transcribed
from published comparison tables, with V-JEPA~2.1 taken from its own
report; neither we nor, for the prior baselines, the cited source
establish a common code-level rerun. This is why we describe it as
reported-protocol aligned rather than verified, and why we do not
extend it with rows from the other group. It also sharpens Ch6: a
backbone-aware protocol is of limited use if the community
reporting the metric has not converged on level~(ii).

The main message of Tables~\ref{tab:ego4d} and~\ref{tab:epic} is
not that one method is universally superior to another, but that
representative numbers remain interpretable only within tightly
matched protocol settings. In particular, benchmark version,
evaluation subset, metric definition, and decoder setup can
materially change the meaning of a reported result. Supplementary Table~S2 contains the full comparison matrix. The main
text retains only the clearest representative rows.
\section{Efficiency and Streaming Deployment}
\label{sec:deploy}

Practical deployment of language-augmented anticipation systems depends
on predictive quality and on whether the system can operate under
realistic compute, memory, and latency constraints.
Except where explicitly stated, the E2/E3 studies are included as transferable deployment mechanisms.
They do not provide direct evidence of anticipation-performance gains; Table~\ref{tab:e23} records this
relation for each method.

We therefore distinguish two cross-cutting deployment dimensions:
\textbf{E2}, which concerns computational and memory efficiency
over long video contexts, and \textbf{E3}, which concerns
streaming and real-time responsiveness. Unlike C1--C4, these are
not functional components of the anticipation pipeline itself, but
deployment-oriented constraints that interact with all four.

\subsection{Efficiency (E2)}

The central challenge in E2 is to reduce computational and memory
cost without discarding the visual or temporal cues that remain
necessary for anticipation. Existing methods do this mainly in two
ways: by reducing the number of tokens that must be processed, or
by controlling how long-context information is retained across
time.

\textbf{Token sparsification.}
SparseVLM~\citep{sparsevlm} suppresses low-value visual tokens
through text--vision attention, PruneVid~\citep{prunevid} removes
static or spatially redundant tokens, and VTS~\citep{vts}
combines key-frame selection with saliency and novelty scoring.
These methods form a lightweight E2 family. They are largely
training-free, can be composed with downstream C1--C4 designs, and are
useful when visual token volume is the main bottleneck.

\textbf{KV-cache and memory management.}
A second E2 family addresses the growth of retained context over
time. AdaCM$^2$~\citep{adacm2} uses a Q-Former-style bridge, introduced as a frozen vision--language interface in BLIP-2~\citep{li2023blip2}, to
retain query-relevant tokens in rolling memory, reducing context
growth toward a near-constant retained budget. ReKV~\citep{di2025rekv}
encodes frames with sliding-window attention and recalls per-frame
KV states at query time, making latency substantially less
sensitive to raw video length. E2 reduces token volume and determines which parts of the past remain
available to the model.

\textbf{Long-context architecture.}
A third E2 direction reduces the cost of the temporal model itself. CLAM~\citep{zhong2026clam} reformulates
cross-attention as linear attention so that retrieval from long-term
memory has linear compute and near-constant memory in the length of
the observation window, and pairs it with a non-autoregressive
decoder that emits the future actions in one shot. The design point
is worth noting alongside the LLM-based decoders reviewed in C3:
where autoregressive decoding makes horizon length a latency
multiplier, one-shot decoding does not, at the cost of losing
explicit conditioning of each predicted action on its predecessors.

\textbf{Edge and wearable deployment.}
The deployment gap becomes most visible on edge hardware. Wearable and edge platforms operate under substantially tighter
memory, energy, and latency budgets than server-class LLM systems,
whereas the weights alone of a 7B-parameter language model occupy
roughly 14\,GB at FP16, before the KV cache, activations, and visual
encoder are accounted for. The total depends on sequence length,
batch size, and quantisation, and we give the weight figure only as an order-of-magnitude anchor, not as a
device-specific deployment estimate. E2 techniques narrow this gap but do not close it. The
most plausible paths toward edge-deployable LLM-augmented
anticipation currently combine aggressive quantisation or
distillation of the language component with event-gated
invocation, in which the LLM is consulted only at task-relevant
moments rather than on every frame, as exemplified by the
streaming designs discussed next. Until such combinations are
demonstrated on anticipation benchmarks, the specific LLM-augmented
anticipation systems reviewed here, which are predominantly built on
unquantised 7B-class decoders evaluated offline, are best regarded as
server-class. We stop short of a general claim: model size,
quantisation, distillation, batching, and API-versus-local execution
vary enough that ``LLM-augmented'' does not by itself determine a
device class. The reporting guidelines in Box~3 accordingly ask
authors to state the device class and quantisation regime explicitly
rather than leaving it to be inferred.

\subsection{Streaming Deployment (E3)}

E3 systems must be efficient on average and respond in a temporally
meaningful way under continuous video input. Existing approaches
mainly improve streaming behaviour either by reducing unnecessary
LLM invocation or by compressing long-term memory so that online
reasoning remains stable over time.

\textbf{VideoLLM-online~\citep{chen2024videollm}.}
VideoLLM-online trains the model to remain silent by default
through Streaming-EOS prediction, invoking language output only
when needed. Its main importance for E3 is conceptual: streaming
anticipation should not be treated as repeated offline inference
on sliding windows, but as a selective online reasoning problem.

\textbf{STREAMMIND~\citep{ding2025streammind}.}
STREAMMIND uses a Cognition Gate to trigger the LLM only on
query-relevant events detected by an upstream extractor. This
makes it a representative event-gated E3 design, in which the
system's responsiveness depends on recognising when reasoning is
actually warranted rather than processing every frame uniformly.

\textbf{MA-LMM~\citep{he2024malmm}.}
MA-LMM addresses E3 from the memory side by compressing long-term
video context into a near-constant-length representation that can
be used with off-the-shelf multimodal LLMs. Its contribution is
therefore complementary to event gating: instead of reducing when
the model reasons, it reduces how much historical state must be
carried forward.

\begin{protocolbox}
\textbf{Reporting for streaming (E3).}
Alongside offline ED or seq-F1, streaming systems should report
FPS, end-to-end latency, LLM call rate, KV-cache budget,
timing-alignment metrics (e.g.\ TimeDiff or Fluency), memory
footprint, and any gating thresholds used to suppress or trigger
reasoning. Server-class and edge-class settings should also be
distinguished explicitly.
\end{protocolbox}

\subsection{Integration of E2/E3 with the Four Components}
\label{sec:e23-integration}

Although E2 and E3 are deployment-oriented constraints, not functional
components, they interact directly with C1--C4.

\textbf{C1 (context construction).}
Aggressive pruning in E2 can easily remove precisely the local
cues on which context quality depends, such as hand--object
interaction regions, narration alignment, or short but informative
state changes. Efficiency mechanisms are therefore safest when
they preserve a minimum task-relevant token budget for salient
event regions.

\textbf{C2 (goal and intention modelling).}
Compression can also become goal-dependent. In methods such as
AdaCM$^2$, the inferred or queried objective helps determine which
visual evidence is worth retaining, creating a feedback loop
between intention modelling and memory management.

\textbf{C3 (future prediction and decoding).}
Streaming mechanisms often act by controlling when decoding is
invoked at all. Streaming-EOS in VideoLLM-online and event-gated
reasoning in STREAMMIND both reduce unnecessary calls to the
decoder, effectively trading continuous responsiveness against
per-invocation accuracy and computational budget.

\textbf{C4 (grounding and executability).}
Grounding can either help or hinder deployment, depending on the
cost of the added feasibility check. Lightweight symbolic or
graph-based constraints may prune invalid candidates early at low
cost, whereas more elaborate geometric, simulator-based, or
world-model validation can itself become a deployment bottleneck.
We would expect grounding overhead to matter most on constrained
hardware, which would favour symbolic over neural feasibility checks
in that regime; we state this as a design hypothesis, since the
surveyed papers do not report the latency measurements that would
settle it. The practical value of C4 in
deployment therefore depends on how cheaply feasibility can be
checked relative to candidate generation.
\section{Cross-Component Discussion}
\label{sec:discussion}

Having reviewed the four functional components individually, we
now synthesise the main findings across the full anticipation
pipeline.

\subsection{Operationalising Goal Ambiguity}
\label{sec:goal-ambiguity-def}

Throughout this survey, ``goal ambiguity'' is used as a
hypothesis-generating moderator, not as an observed dataset variable distinguishing settings where LLM augmentation
tends to add value (e.g., Ego4D-LTA) from settings where it
is hypothesised to contribute less (e.g., short-horizon EK-100). We use the term
as a qualitative proxy; it is not measured in the present synthesis, and the dataset-level classifications offered
elsewhere in this survey (Table~\ref{tab:llm-value},
Fig.~\ref{fig:decision-guide}) are derived from horizon length and
task structure and not from a direct measurement of ambiguity
itself.

A more principled operationalisation would define goal ambiguity
as the normalised entropy of the empirical next-action distribution
over held-out sequences conditioned on a fixed observed prefix, or
as the proportion of prefix-conditioned sequences for which
multiple distinct high-level goal categories remain plausible under
manual annotation~\citep{zhao2024antgpt}. Neither measure is
computed in the present review, because no dataset in our synthesis
set reports it directly. We therefore label all high/low ambiguity entries as heuristic
proxies and treat them as an interpretive device
grounded in task horizon and procedural branching factor, not as a
validated construct, and Fig.~\ref{fig:decision-guide} should
accordingly be read as a post-hoc diagnostic summary of where LLM
augmentation has helped in the reviewed literature. It is not a
prospective selector for a new setting unless ambiguity is first
estimated in that setting.

\subsection{When Do LLMs Add Value?}

\begin{figure}[t]
\centering
\resizebox{0.96\columnwidth}{!}{%
\begin{tikzpicture}[
  font=\sffamily,
  >=Stealth,
  branch/.style={rectangle, rounded corners=6pt, draw=black!55,
    thick, fill=gray!7, text width=4.0cm, align=center,
    minimum height=1.15cm, inner sep=4pt},
  decision/.style={diamond, draw=black!55, thick, fill=olive!10,
    aspect=2.0, text width=2.8cm, align=center, inner sep=2pt,
    minimum height=1.6cm},
  action/.style={rectangle, rounded corners=6pt, draw=cyan!60!black,
    thick, fill=cyan!8, text width=4.1cm, align=center,
    minimum height=1.15cm, inner sep=4pt},
  emerging/.style={rectangle, rounded corners=6pt, draw=orange!70!black,
    thick, fill=orange!10, text width=4.1cm, align=center,
    minimum height=1.15cm, inner sep=4pt},
  lab/.style={font=\scriptsize}
]
\node[decision] (task) at (0,0) {Task\\formulation?};
\node[branch] (single) at (-5.4,-2.6)
  {Single-action, closed-vocabulary anticipation\\report $\tau_a$ and subset};
\node[branch] (sequence) at (5.4,-2.6)
  {Sequence anticipation\\report $Z$, $N$, $K$, and version};
\node[action] (c1) at (-5.4,-5.3)
  {Use a strong C1 backbone as a required diagnostic; treat current EK-100 evidence as Level~(i)};
\node[decision] (amb) at (5.4,-5.3)
  {Multiple plausible goals empirically supported?};
\node[action] (c23) at (9.9,-8.2)
  {Evaluate C2 goal conditioning with stable C3 decoding};
\node[action] (base) at (1.0,-8.2)
  {Benchmark C1+C3; label goal ambiguity as heuristic if unmeasured};
\node[decision] (exec) at (0,-11.0)
  {Executability or feasibility required?};
\node[emerging] (c4) at (5.4,-13.6)
  {Evaluate C4 grounding; evidence remains emerging and domain-specific};
\node[action] (report) at (-5.4,-13.6)
  {Report protocol, uncertainty, calibration, and deployment cost};

\draw[->, thick] (task.west) -| node[lab, near start, above] {single action} (single.north);
\draw[->, thick] (task.east) -| node[lab, near start, above] {sequence} (sequence.north);
\draw[->, thick] (single) -- (c1);
\draw[->, thick] (sequence) -- (amb);
\draw[->, thick] (amb.east) -| node[lab, near start, above] {yes/high} (c23.north);
\draw[->, thick] (amb.west) -| node[lab, near start, above] {no/unknown} (base.north);
\draw[->, thick] (c1.south) |- (exec.west);
\draw[->, thick] (base.south) |- (exec.west);
\draw[->, thick] (c23.south) |- (exec.east);
\draw[->, thick] (exec.east) -| node[lab, near start, above] {yes} (c4.north);
\draw[->, thick] (exec.west) -| node[lab, near start, above] {no} (report.north);
\end{tikzpicture}%
}
\caption{Hypothesis-generating evaluation map with separate branches
for single-action anticipation (parameterised by anticipation time
$\tau_a$) and sequence anticipation (parameterised by horizon $Z$ and
associated protocol axes). The map summarises where the reviewed
evidence suggests testing C1, C2--C3, or C4; it is not a validated
prospective selector. In particular, goal ambiguity is unmeasured in
the underlying benchmarks and must be labelled as a heuristic unless
estimated explicitly (Section~\ref{sec:goal-ambiguity-def}).}
\label{fig:decision-guide}
\end{figure}
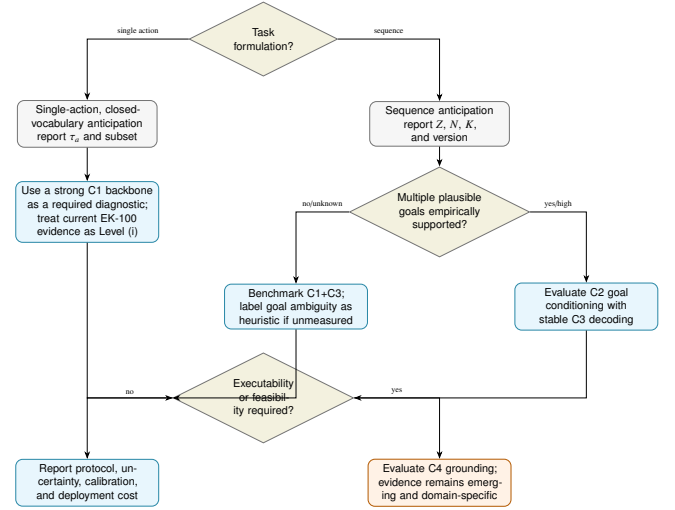

We treat this question as a controlled-comparison problem, not as a
preference between model families. Fig.~\ref{fig:decision-guide} first
separates single-action and sequence formulations, then identifies the
counterfactuals needed before attributing a gain to language-related context,
goal conditioning, decoding, or feasibility checking. In particular, horizon
is not used as a stand-alone selector: task regime, empirically supported goal
branching, representation strength, and the evaluation target must be reported
together. The map is therefore hypothesis-generating and diagnostic, not a
validated prescription for choosing an architecture. Table~\ref{tab:llm-value}
summarises the corresponding protocol-aware evidence and its remaining
confounds.

\begin{table*}[t]
\scriptsize\centering
\caption{Conditions under which language-related components have shown gains,
together with the unresolved short-horizon EK-100 reporting incompatibility.
The incompatibility row is retained to audit comparability, not to rank model
families. Evidence
profiles list publication maturity, component isolation, and protocol
relevance; they are not scalar quality grades. Rows are not matched
causal comparisons unless explicitly stated. LEAP uses a non-MT5R
metric, and the EK-100 row is source-described, unreproduced, and not
pretraining-matched.}
\label{tab:llm-value}
\setlength{\tabcolsep}{3.8pt}
\renewcommand{\arraystretch}{1.25}
\begin{tabularx}{0.98\textwidth}{P{1.85cm}P{1.05cm}P{1.45cm}P{1.50cm}P{1.45cm}YP{2.45cm}P{1.55cm}}
\toprule
\textbf{Setting} &
  \textbf{Horizon} &
  \shortstack[l]{\textbf{Ambiguity}\\\textbf{proxy}} &
  \textbf{Constraint} &
  \textbf{Backbone} &
  \textbf{Evidence} &
  \shortstack[l]{\textbf{Evidence}\\\textbf{profile}} &
  \textbf{Source} \\
\midrule
LTA, goal-ambiguous &
  $Z\!\geq\!8$ & High & Low & Moderate &
  AntGPT (Ego4D-v1 val): inferred-goal conditioning reduces
  verb/noun ED from 0.735/0.753 to 0.724/0.744; increasing
  $N_{\mathrm{seg}}$ from 1 to 8 reduces ED from
  0.734/0.748 to 0.707/0.719 &
  Archival; dedicated; direct & \citep{zhao2024antgpt} \\[2pt]

Procedural + constraints &
  Any & Med. & High & Any &
  LEAP: action-anticipation accuracy on EPIC-KITCHENS
  14.64\%$\to$16.98\% (author ablation, not MT5R) &
  Preprint; partial; metric-mismatched & \citep{dessalene2023leap} \\[2pt]

Rare/long-tail verbs &
  Any & High & Low & Any &
  M-CAT (EK-100 val): adding action/object text raises action
  Recall@5 from 18.4 to 23.7 overall and from 16.0 to 21.0
  on tail classes (author ablation) &
  Preprint; dedicated; direct & \citep{beedu2024mcat} \\[2pt]

Short-horizon, scaled backbone &
  $\tau_a{=}1$s & Low & Low & High &
  Unresolved reporting incompatibility in this Level-(i) matrix: PlausiVL (8B) 27.6,
  V-JEPA~2 39.7, and V-JEPA~2.1 40.8 action MT5R on EK-100 val &
  Mixed maturity; cross-paper; Level~(i) & \citep{assran2025vjepa2,murLabadia2026vjepa21,mittal2024plausivl} \\[2pt]

Streaming, tight latency &
  Any & Any & Low & Any &
  STREAMMIND reports up to 100~FPS in its stated A100 setup; input resolution, batching, output regime, and end-to-end anticipation equivalence are not established here &
  Preprint; no anticipation isolation; transfer & \citep{ding2025streammind} \\
\bottomrule
\end{tabularx}
\par\vspace{2pt}
\footnotesize\textit{Rows are not mutually exclusive. Full protocol metadata are provided in Supplementary Table~S2.}
\end{table*}

\subsection{Evidence Levels and Conditional Investment}

Three patterns emerge, with different levels of evidential maturity.
First, archival author-controlled ablations support the direction of
benefit from goal conditioning and richer observed context in
Ego4D-LTA, although they do not isolate language-model use from all
other design choices~\citep{zhao2024antgpt}. Second, emerging C4
studies suggest that explicit feasibility constraints can improve
planning-oriented outcomes, but this evidence is less mature and often
uses non-standard anticipation metrics~\citep{dessalene2023leap}.
Third, the short-horizon EK-100 record supports no cross-family
performance ordering. Published values under nominally similar labels
are widely separated, while evaluator code, provenance, pretraining,
probe design, loss, and anticipation-time sampling are not jointly
matched. Table~\ref{tab:epic}, Supplementary Table~S2B, and
\texttt{table13\_reporting\_metadata.csv} therefore document a
reporting incompatibility and do not form a leaderboard.

This incompatibility does not establish that non-LLM pretraining is superior,
that language augmentation is unnecessary, or that the rows share a
common evaluator and training recipe. It also cannot locate a
cross-over horizon: no identified study holds backbone and pretraining
fixed while varying language integration across task regimes. EK-100
and Ego4D-LTA further differ in target, output space, metric, and
uncertainty structure, so they should not be interpreted as two points
on a single horizon axis. These unresolved comparisons motivate the
BCAP-Swappable protocol in Section~\ref{sec:challenges}.

\subsection{Three Hypotheses Requiring Controlled Validation}
\label{sec:testable-hypotheses}

The present evidence supports three falsifiable hypotheses, not causal
conclusions.

\begin{enumerate}
\item \textbf{H1: goal-branching hypothesis.} The incremental value of
C2 language-derived goal conditioning increases when multiple distinct
goals remain plausible after the observed prefix. Horizon is an
imperfect proxy; the hypothesis should be tested against an explicit
measure of goal branching or conditional entropy.

\item \textbf{H2: representation-sufficiency hypothesis.} The marginal
benefit of language augmentation decreases when a task is closed-set and
a strong C1 representation already captures the predictive evidence.
This is not established by the current EK-100 record, which remains
confounded by pretraining, probe, loss, and evaluator differences.

\item \textbf{H3: feasibility-value hypothesis.} C4 grounding improves
downstream utility when the evaluation rewards solvability, physical
feasibility, or safety, but may reduce coverage and increase cost under
label-only benchmarks.
\end{enumerate}

\noindent A valid test of these hypotheses requires the
BCAP-Swappable controls of Section~\ref{sec:challenges}; cross-dataset
comparisons or native-system rankings are insufficient.

\subsection{Cross-Component Failure Diagnostics}

The four components do not fail independently. Several recurring
error patterns only become visible when the pipeline is analysed
across component boundaries.

\textbf{C1$\rightarrow$C2: contextual noise propagation.}
When C1 produces semantically weak context, for example through
noisy recogniser output, overly short histories, or impoverished
textual summaries, C2 must rely more heavily on language priors
than on video-grounded evidence. The likely symptom is that adding
goal conditioning does not help, or even degrades performance. In
such cases, the first diagnostic target should be C1 quality
before changing C2 prompting.

\textbf{C2$\rightarrow$C3: goal ambiguity leakage.}
Most current methods effectively use MAP goal inference rather
than propagating uncertainty over plausible goals. The result is a
decoder that may produce sequences strongly consistent with one
goal hypothesis while remaining brittle to slight changes in
context. Poor calibration at long horizons can therefore indicate that the
C2--C3 interface is the main failure point, not C3 in isolation.

\textbf{C3: autoregressive collapse.}
Without explicit repetition control or planning structure,
decoders can collapse toward a small set of high-frequency
actions, especially at longer horizons. PlausiVL's gains from anti-repetition control show that decoding
failures depend on sequence-stability mechanisms as well as model
capacity~\citep{mittal2024plausivl}. The diagnostic signature is
high verb recall paired with low action recall, indicating that
the decoder has learned that frequent verbs are a safe bet.

\textbf{C4: schema and subgoal brittleness.}
Grounding methods can fail when the external structure they depend
on is itself fragile. In symbolic systems, this appears as schema
or ontology brittleness under domain shift; in hierarchical world
models, it appears as semantically plausible but unreachable
subgoals. For such systems, plan solvability and logic
consistency should be measured alongside standard anticipation
metrics.

\textbf{C4 as a deployment trade-off.}
Grounding also affects deployment cost. Lightweight symbolic
checks may sometimes prune invalid candidates early and reduce
wasted decoding effort, whereas simulator-based or world-model
validation may itself become a latency bottleneck. The practical
value of C4 therefore depends on whether feasibility checking is
cheaper than the errors it prevents.

\subsection{The Anticipation--Planning Boundary}

Several recent methods blur the boundary between predicting future
action labels and generating executable plans. VideoPlan treats
anticipation as goal-conditioned future generation; Anticipate
\& Act converts predicted intentions into PDDL goals; and HWM
moves further toward embodied planning by producing subgoals that
function both as anticipation outputs and as control waypoints.
This boundary is conceptually important because it changes what it
means for a predicted future to be ``correct'': sequence agreement
may no longer be sufficient once the output is expected to be
solvable or executable.

A natural implication is that future benchmarks should evaluate
both predictive accuracy and executability. Label-level agreement alone
does not capture downstream usefulness.

\subsection{Concluding Synthesis}
\label{sec:concluding-synthesis}

Table~\ref{tab:design-guide} condenses the practical guidance of
this survey into a single decision table mapping problem regimes
to the components most worth strengthening. It should be read
together with the caveats of
Section~\ref{sec:goal-ambiguity-def}: the mapping reflects the
empirical patterns observed in the reviewed literature under the
stated evidence profiles, not a validated prospective instrument.

\begin{table}[t]
\footnotesize\centering
\caption{Practical design guide: which component to prioritise by
problem regime. The table synthesises the evidence of
Sections~\ref{sec:c1}--\ref{sec:deploy} and inherits their
evidence-profile caveats.}
\label{tab:design-guide}
\setlength{\tabcolsep}{4pt}
\renewcommand{\arraystretch}{1.3}
\begin{tabular}{p{3.3cm}p{4.3cm}}
\toprule
\textbf{Problem regime} & \textbf{Primary component(s)} \\
\midrule
Short-horizon, closed vocabulary ($\tau_a \approx 1$s)
  & A scaled C1 backbone should be evaluated as a required diagnostic baseline before attributing gains to language augmentation \\
Long-horizon, goal-ambiguous ($Z \geq 8$)
  & Evaluate C2 + C3; goal ambiguity is currently a heuristic proxy \\
Safety- or executability-critical
  & Evaluate C4 grounding; evidence remains emerging and domain-specific \\
Real-time / edge deployment
  & E2 + E3: token pruning with event-gated streaming \\
Long-tail and compositional generalisation
  & C1 + C2: open-vocabulary context with structured goals \\
Calibration-critical applications
  & C3 with post-hoc calibration (BCAP Step~6) \\
\bottomrule
\end{tabular}
\end{table}

\begin{table*}[t]
\footnotesize\centering
\caption{Direct answers to the three organising questions. Confidence
reflects the maturity and comparability of the retained evidence, not
a statistical posterior.}
\label{tab:questions-answered}
\setlength{\tabcolsep}{4pt}
\renewcommand{\arraystretch}{1.2}
\begin{tabularx}{\textwidth}{p{2.5cm}Yp{2.2cm}p{4.0cm}}
\toprule
\textbf{Question} & \textbf{Evidence-supported answer} &
\textbf{Confidence} & \textbf{Principal remaining gap} \\
\midrule
Q1: What components define the design space?
& C1--C3 form the core analytical pipeline; C4 is a useful but
less mature grounding/executability extension. The decomposition is
non-unique and methods may span components.
& Moderate--high
& Primary assignment is non-unique and was not blindly recoded in
full by an independent second coder; row-level coding and the
alternate-assignment sensitivity analysis are supplied for audit. \\
Q2: Which design factors matter?
& Context fidelity, goal conditioning, decoding stability, and
feasibility constraints recur across methods. Their relative value
depends on horizon, task structure, and deployment budget.
& Moderate
& Few studies isolate one factor while holding backbone, data, and
compute fixed. \\
Q3: When do language components add value beyond a backbone?
& Author-controlled long-horizon studies support some language-related
components. For EK-100, published tables contain a large unresolved
discrepancy under nominally similar labels, so the present record
cannot support a cross-family performance ordering.
& Low--moderate
& No identified study performs a pretraining-matched, code-verified
comparison across horizons. \\
\bottomrule
\end{tabularx}
\end{table*}

\begin{table*}[t]
\footnotesize\centering
\caption{Claim--evidence map for the main synthesis statements.
``Shared evaluator'' requires demonstrably identical released code,
not merely the same metric name. Confidence is an interpretive evidence
judgement, not a meta-analytic effect size.}
\label{tab:claim-evidence}
\setlength{\tabcolsep}{4pt}
\renewcommand{\arraystretch}{1.22}
\begin{tabular*}{\textwidth}{@{\extracolsep{\fill}}P{4.2cm}P{3.0cm}P{1.35cm}P{1.85cm}P{2.45cm}P{2.35cm}@{}}
\toprule
\textbf{Claim} & \textbf{Principal studies} & \textbf{Groups} &
\textbf{Shared evaluator} & \textbf{Controlled evidence} &
\textbf{Confidence} \\
\midrule
Goal/intention conditioning can improve long-horizon anticipation &
AntGPT, ICVL, INSIGHT, GP-AMS & 4 & No across papers & Dedicated or
partial author ablations & Moderate \\
Semantic/action-history context can replace part of dense video in
selected procedural settings & TransFusion, AAG/AAG+, PALM & 3 & No
across papers & Modality/history ablations & Moderate, task-specific \\
A geometric intent signal can outperform textual conditioning in the
studied regime & TrajPilot & 1 & Within study & Direct conditioning
contrast & Moderate within study; unreplicated \\
Anti-repetition or structured decoding can improve sequence stability
& PlausiVL, VideoPlan, AGA & 3 & No across papers & Heterogeneous author
ablations & Low--moderate \\
Published EK-100 numbers are difficult to reconcile & PlausiVL,
V-JEPA~2/2.1 and transcribed baselines & Multiple & No & No matched
cross-family ablation & High for incompatibility; low for ranking \\
Grounding/verification may improve feasibility & LEAP, FactCheck,
SymAnt and boundary cases & Multiple & No & Mainly within-study or
non-benchmark evidence & Low/emerging \\
Calibration/selective prediction require explicit measurement & Du et
al.; calibration literature & 1 anticipation group & Within
study only & Evaluation protocol, not architecture ablation & Emerging \\
\bottomrule
\end{tabular*}
\end{table*}

\section{Open Challenges}
\label{sec:challenges}

Despite rapid progress, several fundamental challenges remain open
for language-augmented action anticipation. Some concern modelling
capacity, others concern evaluation practice, and several lie at
the boundary between anticipation and planning.

\textbf{Ch1: Compositional generalisation.}
Current methods often learn recurring action co-occurrence
patterns, yet still struggle to compose novel procedure sequences
from previously unseen combinations of familiar components.
Progress will require benchmarks that explicitly test
cross-procedure transfer, as well as models that represent
procedure steps as composable units rather than as opaque sequence
tokens.

\textbf{Ch2: Calibration and uncertainty.}
The temporal-prefix study of Du et~al.~\citep{du2026uncertainty}
directly evaluates calibration, selective prediction, and confidence
reliability for VLM-based early action anticipation on EK-100 and
EGTEA Gaze+, providing emerging evaluation evidence rather than a new
forecasting architecture. 
Most systems output a single ranked list of actions without
reliable probability estimates. Modern neural networks are often
miscalibrated~\citep{guo2017calibration}, and this problem
typically worsens under distribution shift~\citep{ovadia2019trust}.
Future work should therefore report uncertainty-aware metrics such
as ECE, Brier score, and negative log-likelihood alongside ED and
seq-F1 (see BCAP Step~6 below for an explicit operationalisation
bridging these metrics to ranked sequence outputs), and should
include risk--coverage analysis for top-$K$ anticipation following
standard selective-prediction
conventions~\citep{geifman2017selective}; calibration methods such
as Dirichlet calibration remain separately relevant for probability
post-processing~\citep{kull2019dirichlet}.

\textbf{Ch3: Continual adaptation.}
Models trained on static benchmark distributions degrade as
activity patterns, tools, environments, and procedures evolve.
Continual anticipation, that is, updating from streaming
observations without catastrophic forgetting while preserving
previously acquired procedural knowledge, remains largely open.

\textbf{Ch4: Native multimodal anticipation.}
Most current systems remain heavily RGB-centric, while modalities
such as audio, gaze, IMU, and physiological signals are treated as
auxiliary additions rather than native inputs to anticipation.
Yet these modalities may encode intention, salience, or upcoming
state changes more directly than vision alone. Building genuinely
multimodal anticipation systems that fuse such streams from the
ground up remains an important open frontier.

\textbf{Ch5: Affordance grounding without manual schemas.}
Grounding-oriented methods such as LEAP and SymAnt suggest that
explicit constraints may improve executability and consistency,
but they still depend on manually curated schemas, ontologies, or
symbolic structures that are expensive to build and brittle under
domain shift. Learning transferable affordance constraints
directly from video, while retaining enough interpretability for
safety-critical deployment, remains unresolved.

\textbf{Ch6: Backbone-aware evaluation.}
A central unresolved issue is that gains attributed to LLM
augmentation are rarely separated from visual backbones, pretraining
corpora, training compute, or protocol drift. The field therefore
lacks a reliable answer to when language integration itself adds
value over a strong non-LLM baseline.

We propose the \textbf{Backbone-Aware Comparison and Ablation
Protocol (BCAP)}. ``Aware'' is deliberate: the Native branch makes
backbone confounds visible but does not control them, whereas only the
Swappable branch supports causal within-family ablation.

\begin{enumerate}\itemsep2pt
  \item \textbf{Declare the branch and claim type.}
  \emph{BCAP-Native} is a descriptive system comparison: each method
  keeps its constitutive encoder, while metric, split, horizon,
  modality budget, and candidate count are aligned as far as possible.
  It supports claims about reported systems, not the causal effect of
  an LLM. \emph{BCAP-Swappable} is a causal within-family comparison:
  a common visual backbone and data pipeline are fixed while only the
  language-related component changes. Architecturally coupled systems
  that cannot accept a shared encoder remain Native-only.

  \item \textbf{Use code-verifiable evaluation.}
  Report Ego4D-LTA ED at the official $Z{=}20$ setting with label-space
  version, $N$, $K$, and evaluation-code repository plus commit hash.
  Report EK-100 action Mean Top-5 Recall (class-mean) on a named split
  at $\tau_a{=}1$s. State whether every Top-5 quantity is class-mean or
  instance-level, and keep overall, unseen-participant, tail, and
  hidden-test subsets separate.

  \item \textbf{Control training data and compute.}
  For Swappable comparisons, match training examples, optimisation
  steps, augmentations, trainable-parameter policy, and compute budget
  where feasible. For both branches, disclose pretraining sources and
  volumes, benchmark overlap, potential data contamination, total and
  trainable parameters, and any extra supervision. Unmatched factors
  must be listed as residual confounds rather than absorbed into an
  LLM-versus-non-LLM claim.

  \item \textbf{Report a horizon sweep.}
  Keep $Z{=}20$ as the official Ego4D-LTA row and add clearly labelled
  BCAP diagnostic rows for $Z\in\{4,8,12\}$. Truncate the ground-truth
  sequence to the first $Z$ actions and normalise diagnostic edit
  distance by $Z$. Do not present diagnostic rows as official challenge
  results or select only the best horizon.

  \item \textbf{Test backbone sensitivity and use a bounded anchor.}
  Repeat Swappable ablations with at least one weaker backbone. Pair
  the common encoder with a specified non-LLM decoder of bounded depth,
  width, and parameter count (for example, a fixed transformer decoder
  or RU-LSTM). Report both decoder parameter counts when exact matching
  is infeasible.

  \item \textbf{Evaluate calibration and selective risk.}
  At step level, report NLL, Brier score, and ECE over fixed verb and
  noun vocabularies. A sequence-level multiclass Brier score is valid
  only for a fixed candidate space guaranteed to contain the ground
  truth. Otherwise define a binary sequence-correctness event and
  report its Brier score, ECE, and risk--coverage. Length-normalise and
  calibrate beam scores on held-out data; do not treat raw beam scores
  as probabilities.

  \item \textbf{Make prompts and stochasticity reproducible.}
  Report the exact LLM/VLM provider or checkpoint, version/date,
  quantisation, prompt templates, system messages, number and selection
  of exemplars, decoding temperature, top-$p$/top-$k$, beam settings,
  random seeds, and label-normalisation rules. For stochastic methods,
  report at least three runs or repeated API evaluations with mean and
  an uncertainty interval.

  \item \textbf{Report deployment cost.}
  State GPU/accelerator type, precision, batch size, peak memory,
  end-to-end latency, throughput, LLM call rate, token counts, and, for
  paid APIs, the pricing date and per-example cost. Separate encoder,
  language-model, and grounding/planning costs where possible.

  \item \textbf{Provide row-level provenance and complete metadata.}
  Every result row must include backbone, modality, label space,
  horizon, subset, decoding policy, training/pretraining metadata,
  source table or figure, evaluation-code commit, and whether the value
  was transcribed, code-verified, or independently reproduced. Release
  configuration files and prompts sufficient to recreate the row.
\end{enumerate}

\noindent
BCAP is a reporting and ablation framework, not a claim that every
architecture can be reduced to one backbone. Its purpose is to
separate transparent system comparison from causal attribution and to
make future evidence cumulative. The nine-item checklist in the
supplementary material maps one-to-one to the nine steps above.

\subsection{Illustrative Retrospective BCAP-Native Audit}
\label{sec:bcap-worked}

Table~\ref{tab:bcap-worked} retrospectively audits a selected
subset of BCAP-Native dimensions for three representative systems;
the complete nine-item audit is provided in Supplementary Table~S7A. This is a documentation audit, not a rerun:
``partial'' means that the paper or public artefact reports some of the
required information but not a version-frozen configuration sufficient
to recreate the survey row under shared code.

\begin{table*}[t]
\footnotesize\centering
\caption{Selected BCAP-Native dimensions for three representative systems. Statuses reflect the
published record and public artefacts inspected for this review; no
row was independently reproduced.}
\label{tab:bcap-worked}
\setlength{\tabcolsep}{4pt}
\renewcommand{\arraystretch}{1.2}
\begin{tabularx}{\textwidth}{p{3.5cm}YYY}
\toprule
\textbf{BCAP item} & \textbf{AntGPT~\citep{zhao2024antgpt}} &
\textbf{PlausiVL~\citep{mittal2024plausivl}} &
\textbf{V-JEPA~2.1~\citep{murLabadia2026vjepa21}} \\
\midrule
Evaluation implementation/version
& Partial: an official repository and LTA inference path were identified;
commit and file locators are recorded in Supplementary Table~S7B, but
the survey row was not rerun.
& No versioned public implementation was identified in this audit; the
status is therefore based on the paper description only.
& Partial: an official repository, EK-100 evaluation script, and model
configuration were identified and frozen in Supplementary Table~S7B;
the reported row was not rerun. \\
Pretraining disclosure
& Partial: constituent checkpoints are named; exposure is not
harmonised with comparison systems.
& Partial: backbone/checkpoint information is reported; data volume is
not matched to the non-LLM encoders.
& Substantial disclosure of model/data scaling, but it remains
unmatched to the language-augmented systems. \\
Prompt or decoding reproducibility
& Partial: prompting strategy and variants are described; a complete
version-frozen run configuration is not available for every row.
& Partial: plausibility and decoding mechanisms are described; complete
cross-paper run settings are not standardised.
& Not applicable to an LLM prompt; attentive-probe and evaluation
configuration still require versioned reporting. \\
Repeated stochastic evaluation
& No uncertainty interval from repeated LLM evaluations identified.
& No BCAP-style repeated-run uncertainty interval identified.
& No repeated-run interval attached to the reported EK-100 survey row. \\
Deployment cost
& Not reported in the full BCAP hardware/memory/token-cost format.
& Not reported in the full BCAP hardware/memory/token-cost format.
& Compute is discussed for the model family, but not as a matched
per-example BCAP deployment row. \\
Permitted inference
& Descriptive system comparison; author ablations support selected
within-system component claims.
& Descriptive system comparison; author ablations support plausibility
and repetition-control claims.
& Descriptive evidence for a densely pretrained C1 system; no causal
claim about removing an LLM from a matched architecture. \\
\bottomrule
\end{tabularx}
\end{table*}

\noindent Under BCAP-Native, these systems can be compared only as
reported systems. The causal effect of language-model integration
cannot be isolated because backbone, pretraining, architecture, and
training budgets are not jointly controlled. The worked example also
shows why public code alone is insufficient unless a result row is
linked to a versioned evaluator and complete configuration.

\section{Limitations of This Review}
\label{sec:limits}
\label{sec:limitations}

Like any review of a rapidly evolving field, the present survey is
subject to methodological, evidential, and coverage limitations.

\textbf{Search methodology and reproducibility.}
The initial retrieval was not prospectively logged at PRISMA-level
granularity, so its complete attrition path cannot be reconstructed. The
prospective verification passes and archived dispositions reduce vocabulary
sensitivity but do not convert the review into an exhaustive census. The final
69-work set should therefore be interpreted as the retained structured
synthesis set under the reported scope; a researcher applying different
boundary rules could obtain a modestly different set. Full retrieval dates,
query status, and candidate dispositions are reported once in
Section~\ref{sec:method} and Supplementary Section~S3 rather than repeated
here.

\textbf{Focal-set classification.}
The condition-satisfaction tallies in Supplementary Table~S1
reflect judgement-based coding of the cited sources. The final 25-row
matrix was not blindly recoded in full by a second author, so no
inter-rater coefficient is reported. To prevent this limitation from
being hidden behind a single grade, the supplementary data expose primary
component, focal group, component isolation, protocol relevance, and
source location row by row; Supplementary Section~S6 also reports an
alternate-assignment sensitivity analysis. These measures make the
coding auditable and show that the principal conclusions do not depend
on one primary-assignment rule, but they are not a substitute for
future independent agreement measurement.

\textbf{Protocol comparability.}
Source-level alignment cannot eliminate differences in split,
annotation version, visual pretraining, modality, probe, loss, context
window, or decoder. Section~\ref{sec:benchmarks} therefore separates
three levels: source-described alignment, shared-code verification,
and independent reproduction. The Ego4D table retains only rows with
complete version and verb/noun/action edit-distance provenance. The
EK-100 table is a reporting-incompatibility audit at the first level, not a
leaderboard: its sources use nominally similar labels, but pretraining,
probe design, evaluator traceability, and rerun status remain
unmatched. Table~\ref{tab:epic} exposes confounding by pretraining
regime, probe design, and evaluator traceability; it does not support a
comparative claim about those regimes or about language-model use. Resolving that
question requires the BCAP-Swappable experiment proposed in
Section~\ref{sec:challenges}, which is not executed in this review.

\textbf{Preprint and recency dependence.}
The review's short-horizon reporting audit includes recent evidence,
including V-JEPA~2 and V-JEPA~2.1, that had not undergone independent
peer review at the time of writing.
These results are therefore treated throughout as emerging
evidence rather than as settled archival findings. More generally,
any review of a literature moving at this rate is dated on the day
it is submitted.

\textbf{Scope and coverage bias.}
The survey is centred on anticipation-focused works with
action-label outputs, especially in egocentric benchmarks such as
Ego4D and EPIC-KITCHENS. Visual-planning, robotics, and
vision-language-action papers are included only when they
directly illuminate anticipation design or executability.
Third-person and procedural benchmarks such as Assembly101, COIN,
CrossTask, 50Salads, and Breakfast are therefore discussed more
selectively than the core egocentric benchmarks, and multilingual
or non-English research threads are likely underrepresented.
A complementary review centred on third-person, industrial, and
robotic settings would therefore be valuable.

\textbf{Taxonomic judgement.}
The component assignment used throughout the survey is
defensible but not unique. Primary assignment was based on the
largest reported marginal gain where ablation evidence was
available, or otherwise on the method's primary stated
contribution. Alternative taxonomic choices would be possible,
especially for multi-component methods that jointly modify context
construction, goal modelling, decoding, and grounding.

\textbf{Screening consistency.}
Title and abstract screening was conducted independently by two of
the three authors, with disagreements resolved by discussion until
consensus. However, no formal inter-rater statistic such as
Cohen's kappa was computed, which limits the extent to which
screening consistency can be quantified beyond raw agreement.
Full-text assessment of borderline records was conducted jointly.
Future systematic reviews in this subfield would benefit from a
pre-registered screening protocol with formal reliability
measurement from the outset.

\textbf{Reporting dependence.}
Several dimensions that are central to deployment and trustworthy
use, including calibration, latency, memory cost, and
executability, remain inconsistently reported across papers. This
limits the strength of cross-paper synthesis along precisely the
axes that matter most for real-world deployment. The
protocol-metadata tables in this survey are intended as one step
toward more standardised reporting, but they cannot compensate for
missing primary-study documentation.

\section{Conclusion}
\label{sec:conclusion}

The central contribution of this survey is an evidence-aware design map
that jointly locates the task regime and the functional insertion point
of language-derived information. C1--C4 is therefore used not merely to
name stages, but to identify the counterfactual, failure diagnosis, and
evidence claim appropriate to context construction, goal/intention
modelling, future decoding, and an adjacent emerging grounding/
executability extension. The map is applied to 25 focal methods within
a retained synthesis set of 69 works, with 10 additional evidence
records reported separately; it is not presented as an exhaustive
systematic census.

The protocol audit supports a narrower conclusion than a conventional
leaderboard. Author-controlled studies provide evidence that semantic
context, explicit goal conditioning, and sequence-stability mechanisms
can help in particular long-horizon settings, but they rarely hold
backbone, pretraining, supervision, decoder, and evaluator fixed. The
EK-100 discrepancy is accordingly a case study in unresolved reporting
incompatibility, not a verdict on language augmentation. C4 provides a
useful vocabulary for feasibility-sensitive systems, while its evidence
base remains heterogeneous and less mature.

We therefore formulate the survey's main synthesis as three hypotheses:
language-derived goals may be most useful when goal branching is
measurably high; their marginal value may shrink when a strong C1
representation is already sufficient for a closed-set task; and C4 may
add value when downstream evaluation rewards feasibility rather than
label agreement alone. Goal ambiguity and horizon are moderators to be
measured and controlled, not explanations established by the present
cross-paper record.

BCAP specifies the matched comparisons needed to test these hypotheses
through versioned evaluation, data and compute disclosure, explicit
prompts/checkpoints, repeated stochastic evaluation, calibration,
deployment cost, and row-level provenance. The accompanying coded
package and versioned catalogue are intended to make the review
extensible while preserving the distinction between auditable evidence
and causal validation.

\section*{Declarations}
Use of AI-assisted tools: The authors utilized AI-powered
tools to check grammar and enhance the academic quality of
the text, which the authors primarily wrote.

\section*{Acknowledgements}
The authors acknowledge institutional support from the University of
Exeter.

\section*{Supplementary Material}

Supplementary material is provided in a separate file
(\texttt{Supplementary\_Material.pdf}) and contains linked Sections
S0--S7, including the benchmark sub-audits S2A and S2B, covering the
closest-survey audit, focal-method inventory, benchmark comparability,
search and verification procedures, multidimensional evidence
assessment, BCAP, taxonomy sensitivity, and machine-readable audit
materials.


\end{document}